\documentclass{article} 
\usepackage{iclr2017_conference,times}
\usepackage{hyperref}
\usepackage{url}

\usepackage{booktabs}       
\usepackage{amsfonts}       
\usepackage{nicefrac}       
\usepackage{microtype}      
\usepackage{subfigure}
\usepackage[pdftex]{graphicx}
\usepackage[ruled,vlined]{algorithm2e}
\usepackage{amsmath}
\usepackage{multirow}

\title{Pruning Filters for Efficient ConvNets}

\author{Hao Li\thanks{Work done at NEC Labs}\\  
  University of Maryland\\  
  \texttt{haoli@cs.umd.edu} \\
  \And
  Asim Kadav \\
  NEC Labs America \\
  \texttt{asim@nec-labs.com} \\
  \And
  Igor Durdanovic \\
  NEC Labs America \\
  \texttt{igord@nec-labs.com} \\
  \And
  Hanan Samet\thanks{Supported in part by the NSF under Grant IIS-13-2079} \\
  University of Maryland\\
  \texttt{hjs@cs.umd.edu} \\
  \And
  Hans Peter Graf \\
  NEC Labs America \\
  \texttt{hpg@nec-labs.com} \\
}

\iclrfinalcopy 

\begin{document}

\maketitle

\begin{abstract}

The success of CNNs in various applications is accompanied by a significant increase in the computation and parameter storage costs. 
Recent efforts toward reducing these overheads involve pruning and compressing the weights of various layers without hurting original accuracy. 
However, magnitude-based pruning of weights reduces a significant number of parameters from the fully connected layers and may not adequately reduce the computation costs in the convolutional layers due to irregular sparsity in the pruned networks.
We present an acceleration method for CNNs, where we prune filters from CNNs that are identified as having a small effect on the output accuracy. 
By removing whole filters in the network together with their connecting feature maps, the computation costs are reduced significantly.
In contrast to pruning weights, this approach does not result in sparse connectivity patterns. 
Hence, it does not need the support of sparse convolution libraries and can work with existing efficient BLAS libraries for dense matrix multiplications. 
We show that even simple filter pruning techniques can reduce inference costs for VGG-16 by up to 34\% and ResNet-110 by up to 38\% on CIFAR10 while regaining close to the original accuracy by retraining the networks.  

\end{abstract}
\section{Introduction}

The ImageNet challenge has led to significant advancements in exploring various architectural choices in CNNs~(\cite{ILSVRC15,alexnet,vgg,googlenet,resnet}). 
The general trend since the past few years has been that the networks have grown deeper, with an overall increase in the number of parameters and convolution operations. 
These high capacity networks have significant inference costs especially when used with embedded sensors or mobile devices where computational and power resources may be limited.
For these applications, in addition to accuracy, computational efficiency and small network sizes are crucial enabling factors~(\cite{inceptionv3}).
In addition, for web services that provide image search and image classification APIs that operate on a time budget often serving hundreds of thousands of images per second, benefit significantly from lower inference times.

There has been a significant amount of work on reducing the storage and computation costs by model compression  (\cite{obd, OBS, srinivas2015data,han2015learning, mariet2015diversity}).
Recently~\cite{han2015learning, deepcompression} report impressive compression rates on AlexNet (\cite{alexnet}) and VGGNet (\cite{vgg}) by pruning weights with small magnitudes and then retraining without hurting the overall accuracy. 
However, pruning parameters does not necessarily reduce the computation time since the majority of the parameters removed are from the fully connected layers where the computation cost is low, 
e.g., the fully connected layers of VGG-16 occupy 90\% of the total parameters but only contribute less than 1\% of the overall floating point operations (FLOP).
They also demonstrate that the convolutional layers can be compressed and accelerated (\cite{squeezenet}), but additionally require sparse BLAS libraries or even specialized hardware~(\cite{han2016eie}).
Modern libraries that provide speedup using sparse operations over CNNs are often limited~(\cite{googlenet,sparce_cnn}) and maintaining sparse data structures also creates an additional storage overhead which can be significant for low-precision weights. 

Recent work on CNNs have yielded deep architectures with more efficient design~(\cite{googlenet, inceptionv3, he2015convolutional, resnet}), in which the fully connected layers are replaced with average pooling layers~(\cite{nin, resnet}), which reduces the number of parameters significantly.
The computation cost is also reduced by downsampling the image at an early stage to reduce the size of feature maps~(\cite{he2015convolutional}).
Nevertheless, as the networks continue to become deeper, the computation costs of convolutional layers continue to dominate. 

CNNs with large capacity usually have significant redundancy among different filters and feature channels. 
In this work, we focus on reducing the computation cost of well-trained CNNs by pruning filters. 
Compared to pruning weights across the network, filter pruning is a naturally structured way of pruning without introducing sparsity and therefore does not require using sparse libraries or any specialized hardware.
The number of pruned filters correlates directly with acceleration by reducing the number of matrix multiplications, which is easy to tune for a target speedup.
In addition, instead of layer-wise iterative fine-tuning (retraining), we adopt a \emph{one-shot} pruning and retraining strategy to save retraining time for pruning filters across multiple layers, which is critical for pruning very deep networks. 
Finally, we observe that even for ResNets, which have significantly fewer parameters and inference costs than AlexNet or VGGNet, still have about 30\% of FLOP reduction without sacrificing too much accuracy.
We conduct sensitivity analysis for convolutional layers in ResNets that improves the understanding of ResNets. 
\section{Related Work}
The early work by~\cite{obd} introduces Optimal Brain Damage, which prunes weights with a theoretically justified saliency measure. 
Later, \cite{OBS} propose Optimal Brain Surgeon to remove unimportant weights determined by the second-order derivative information.
~\cite{mariet2015diversity} reduce the network redundancy by identifying a subset of diverse neurons that does not require retraining.
However, this method only operates on the fully-connected layers and introduce sparse connections.

To reduce the computation costs of the convolutional layers, past work have proposed to approximate convolutional operations by representing the weight matrix as a low rank product of two smaller matrices without changing the original number of filters~(\cite{denil2013predicting,jaderberg2014speeding,zhang2015efficient,zhang2015accelerating,tai2015convolutional,ioannou2015training}).
Other approaches to reduce the convolutional overheads include using FFT based convolutions
(\cite{mathieu2013fast}) and fast convolution using the Winograd algorithm (\cite{lavin2015fast}).
Additionally, quantization~(\cite{deepcompression}) and binarization~(\cite{xnornet,binarynet}) can be used to reduce the model size and lower the computation overheads.
Our method can be used in addition to these techniques to reduce computation costs without incurring additional overheads.

Several work have studied removing redundant feature maps from a well trained network~(\cite{anwar2015structured,polyak2015channel}).
\cite{anwar2015structured} introduce a three-level pruning of the weights and locate the pruning candidates using particle filtering, which selects the best combination from a number of random generated masks. 
~\cite{polyak2015channel} detect the less frequently activated feature maps with sample input data for face detection applications.
We choose to analyze the filter weights and prune filters with their corresponding feature maps using a simple magnitude based measure, without examining possible combinations.
We also introduce network-wide holistic approaches to prune filters for simple and complex convolutional network architectures.

Concurrently with our work, there is a growing interest in training compact CNNs with sparse constraints~(\cite{group_brain_damage,compactcnn,structured_sparsity_nips16}).
~\cite{group_brain_damage} leverage group-sparsity on the convolutional filters to achieve structured brain damage, i.e., prune the entries of the convolution kernel in a group-wise fashion.
~\cite{compactcnn} add group-sparse regularization on neurons during training to learn compact CNNs with reduced filters.
~\cite{structured_sparsity_nips16} add structured sparsity regularizer on each layer to reduce trivial filters, channels or even layers. 
In the filter-level pruning, all above work use $\ell_{2,1}$-norm as a regularizer.
Similar to the above work, we use $\ell_1$-norm to select unimportant filters and physically prune them.
Our fine-tuning process is the same as the conventional training procedure, without introducing additional regularization.
Our approach does not introduce extra layer-wise meta-parameters for the regularizer except for the percentage of filters to be pruned, which is directly related to the desired speedup.
By employing stage-wise pruning, we can set a single pruning rate for all layers in one stage.

\section{Pruning Filters and Feature Maps}

\begin{figure*}[htbp]
\centering
\begin{tabular}{l}
      \includegraphics[width=0.8\linewidth]{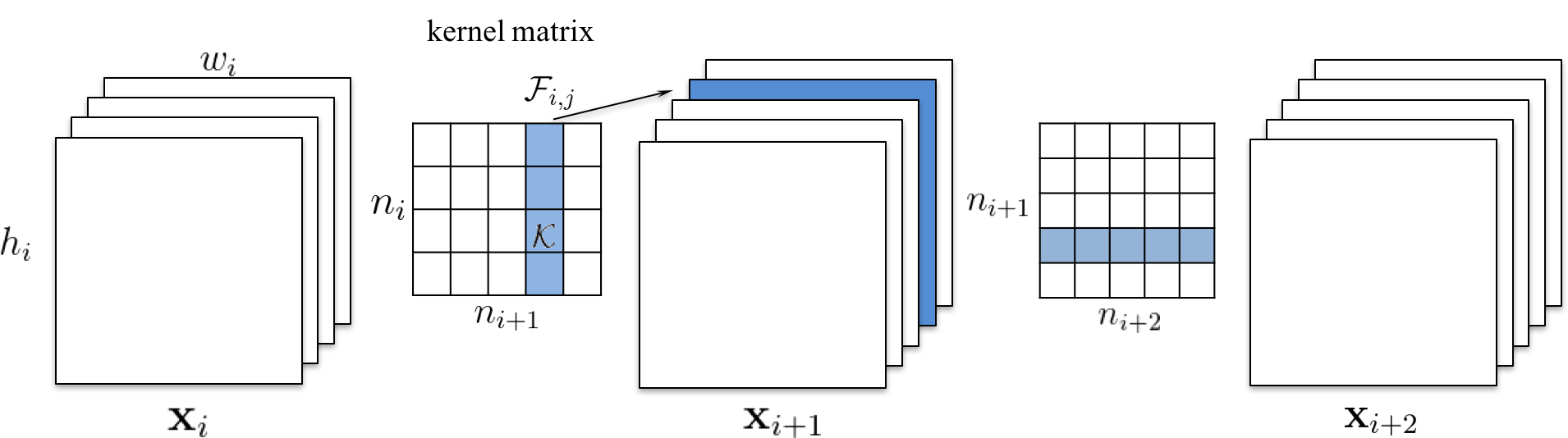}
\end{tabular}
\caption{Pruning a filter results in removal of its corresponding feature map and related kernels in the next layer. 
}
\label{fig:pruning}
\end{figure*}

Let $n_i$ denote the number of input channels for the $i$th convolutional layer and $h_i/w_i$ be the height/width of the input feature maps.
The convolutional layer transforms the input feature maps $\mathbf{x}_{i} \in \mathbb{R}^{n_{i}\times h_i \times w_i}$ into the output feature maps $\mathbf{x}_{i+1} \in \mathbb{R}^{n_{i+1} \times h_{i+1} \times w_{i+1}}$, which are used as input feature maps for the next convolutional layer.
This is achieved by applying $n_{i+1}$ 3D filters $\mathcal{F}_{i, j} \in \mathbb{R}^{n_{i} \times k \times k} $ on the $n_{i}$ input channels, in which one filter generates one feature map.
Each filter is composed by $n_i$ 2D kernels $\mathcal{K} \in \mathbb{R}^{k\times k}$ (e.g., $3 \times 3$).
All the filters, together, constitute the kernel matrix $\mathcal{F}_{i}\in \mathbb{R}^{n_{i} \times n_{i+1} \times k \times k}$.
The number of operations of the convolutional layer is $n_{i+1}n_{i}k^2h_{i+1}w_{i+1}$.
As shown in Figure~\ref{fig:pruning}, when a filter $\mathcal{F}_{i,j}$ is pruned, its corresponding feature map $\mathbf{x}_{{i+1},j}$ is removed, which reduces $n_{i}k^2h_{i+1}w_{i+1}$ operations.
The kernels that apply on the removed feature maps from the filters of the next convolutional layer are also removed,
which saves an additional $n_{i+2}k^2h_{i+2}w_{i+2}$ operations.
Pruning $m$ filters of layer $i$ will reduce $m/n_{i+1}$ of the computation cost for both layers $i$ and $i+1$.

\subsection{Determining which filters to prune within a single layer}
\begin{figure*}[h]
\centering
\begin{tabular}{l}
   \subfigure[Filters are ranked by $s_j$]{
   \includegraphics[width=0.33\linewidth]{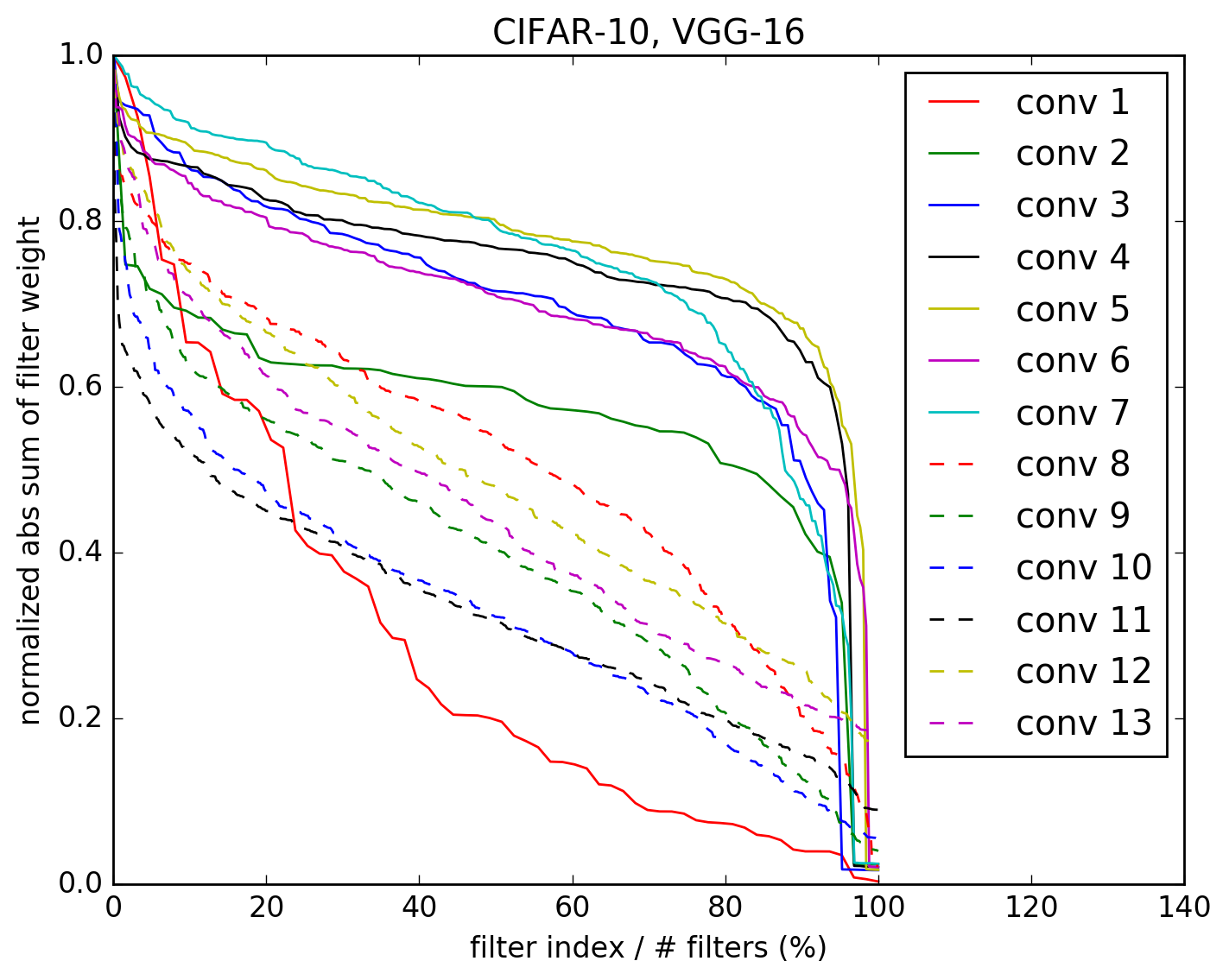}
      \label{fig:weight_sum:a}
   }
   \subfigure[Prune the smallest filters]{
   \includegraphics[width=0.33\linewidth]{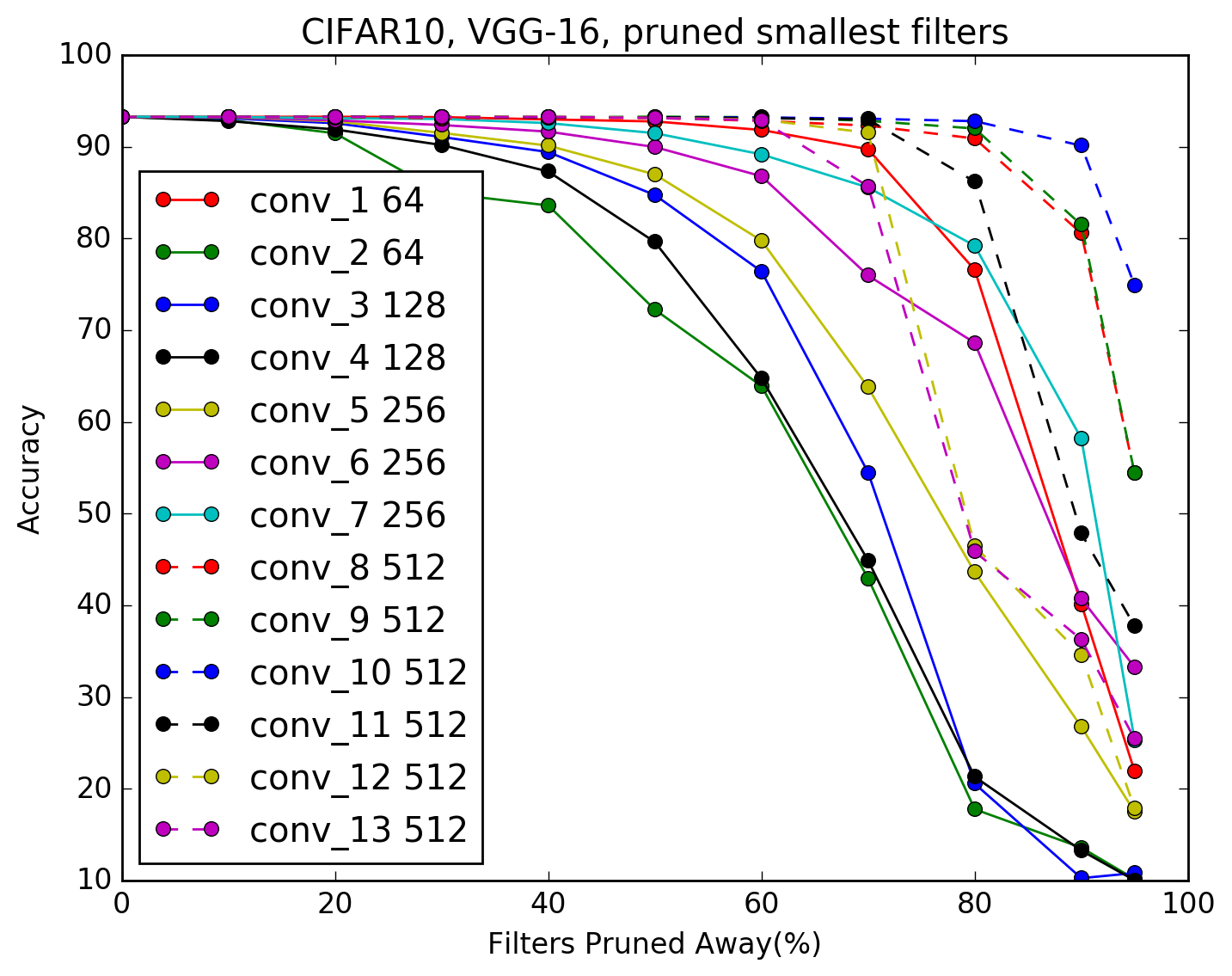}
      \label{fig:prune_smallest_filter}
   }
 \subfigure[Prune and retrain]{
  \includegraphics[width=0.33\linewidth]{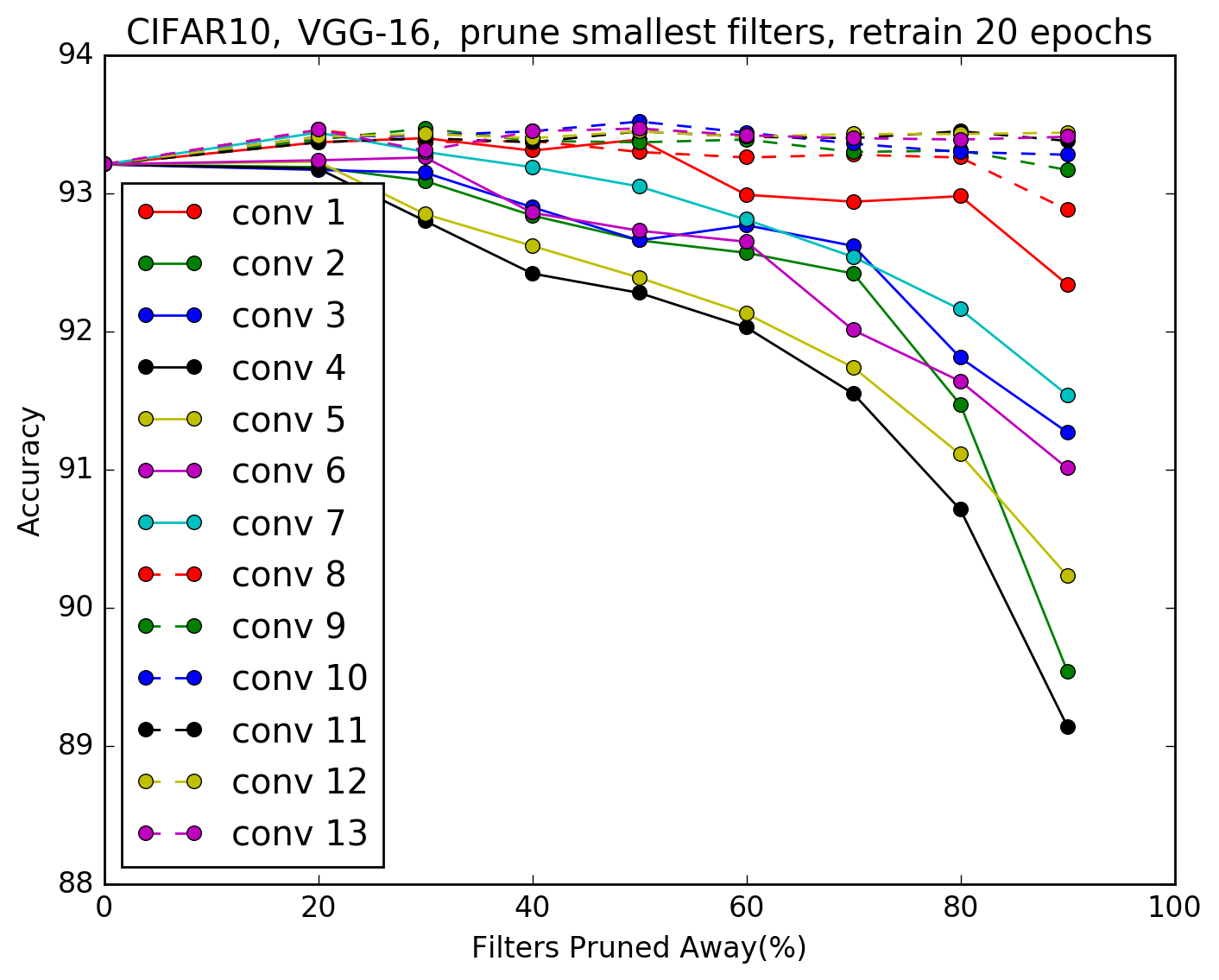}
  \label{fig:retrain_vgg}
  }
\end{tabular}
\caption{(a) Sorting filters by absolute weights sum for each layer of VGG-16 on CIFAR-10.
The x-axis is the filter index divided by the total number of filters.
The y-axis is the filter weight sum divided by the max sum value among filters in that layer.
 (b) Pruning filters with the lowest absolute weights sum and their corresponding test accuracies on CIFAR-10.
 (c) Prune and retrain for each single layer of VGG-16 on CIFAR-10. Some layers are sensitive and it can be harder to recover accuracy after pruning them.}
\label{fig:weight_sum}
\end{figure*}

Our method prunes the less useful filters from a well-trained model for computational efficiency while minimizing the accuracy drop.
We measure the relative importance of a filter in each layer by calculating the sum of its absolute weights $\sum|\mathcal{F}_{i,j}|$, i.e., its $\ell_1$-norm $\|\mathcal{F}_{i,j}\|_1$.
Since the number of input channels, $n_i$, is the same across filters, $\sum|\mathcal{F}_{i,j}|$ also represents the average magnitude of its kernel weights.
This value gives an expectation of the magnitude of the output feature map.
Filters with smaller kernel weights tend to produce feature maps with weak activations as compared to the other filters in that layer.
Figure~\ref{fig:weight_sum:a} illustrates the distribution of filters' absolute weights sum for each convolutional layer in a VGG-16 network trained on the CIFAR-10 dataset, where the distribution varies significantly across layers.
We find that pruning the \emph{smallest} filters works better in comparison with pruning the same number of \emph{random} or \emph{largest} filters (Section~\ref{sec:random_largest_filters}). 
Compared to other criteria for activation-based feature map pruning (Section~\ref{sec:pruning_activation}), we find $\ell_1$-norm is a good criterion for data-free filter selection.

The procedure of pruning $m$ filters from the $i$th convolutional layer is as follows:
\begin{enumerate}
\item For each filter $\mathcal{F}_{i,j}$, calculate the sum of its absolute kernel weights $s_{j} = \sum_{l=1}^{n_i} \sum|\mathcal{K}_l|$.
\vspace{-1mm}
\item Sort the filters by $s_j$.
\vspace{-1mm}
\item Prune $m$ filters with the smallest sum values and their corresponding feature maps. 
The kernels in the next convolutional layer corresponding to the pruned feature maps are also removed.
\vspace{-1mm}
\item A new kernel matrix is created for both the $i$th and $i+1$th layers, and the remaining kernel weights are copied to the new model.
\end{enumerate}

\textbf{Relationship to pruning weights}
Pruning filters with low absolute weights sum is similar to pruning low magnitude weights~(\cite{han2015learning}).
Magnitude-based weight pruning may prune away whole filters when all the kernel weights of a filter are lower than a given threshold. 
However, it requires a careful tuning of the threshold and it is difficult to predict the exact number of filters that will eventually be pruned. 
Furthermore, it generates sparse convolutional kernels which can be hard to accelerate given the lack of efficient sparse libraries, especially for the case of low-sparsity.

\paragraph{Relationship to group-sparse regularization on filters}
Recent work (\cite{compactcnn,structured_sparsity_nips16}) apply group-sparse regularization ($\sum_{j=1}^{n_i}\|\mathcal{F}_{i,j}\|_2$ or $\ell_{2,1}$-norm) on convolutional filters,
which also favor to zero-out filters with small $l_2$-norms, i.e. $\mathcal{F}_{i,j}=\mathbf{0}$.
In practice, we do not observe noticeable difference between the $\ell_2$-norm and the $\ell_1$-norm for filter selection, 
as the important filters tend to have large values for both measures (Appendix~\ref{sec:pruning_filters_l2}).
Zeroing out weights of multiple filters during training has a similar effect to pruning filters with the strategy of iterative pruning and retraining as introduced in Section~\ref{sec:retrain}. 

\subsection{Determining single layer's sensitivity to pruning}

To understand the sensitivity of each layer, we prune each layer independently and evaluate the resulting pruned network's accuracy on the validation set.
Figure~\ref{fig:prune_smallest_filter} shows that layers that maintain their accuracy as filters are pruned away correspond to layers with larger slopes in Figure~\ref{fig:weight_sum:a}.
On the contrary, layers with relatively flat slopes are more sensitive to pruning.
We empirically determine the number of filters to prune for each layer based on their sensitivity to pruning.
For deep networks such as VGG-16 or ResNets, we observe that layers in the same stage (with the same feature map size) have a similar sensitivity to pruning.
To avoid introducing layer-wise meta-parameters, we use the same pruning ratio for all layers in the same stage.
For layers that are sensitive to pruning, we prune a smaller percentage of these layers or completely skip pruning them.

\subsection{Pruning filters across multiple layers}
We now discuss how to prune filters across the network. 
Previous work prunes the weights on a layer by layer basis, followed by iteratively retraining and compensating for any loss of accuracy~(\cite{han2015learning}).
However, understanding how to prune filters of multiple layers at once can be useful:
1) For deep networks, pruning and retraining on a layer by layer basis can be extremely time-consuming
2) Pruning layers across the network gives a holistic view of the robustness of the network resulting in a smaller network
3) For complex networks, a holistic approach may be necessary. For example, for the ResNet, pruning the identity feature maps or the second layer of each residual block results in additional pruning of other layers.

To prune filters across multiple layers, we consider two strategies for layer-wise filter selection:
\begin{itemize}
\item \emph{Independent} pruning determines which filters should be pruned at each layer independent of other layers.
\item \emph{Greedy} pruning accounts for the filters that have been removed in the previous layers.
This strategy does not consider the kernels for the previously pruned feature maps while calculating the sum of absolute weights.
\end{itemize}

\begin{figure*}[htbp]
\centering
\begin{tabular}{l}
      \includegraphics[width=0.45\linewidth]{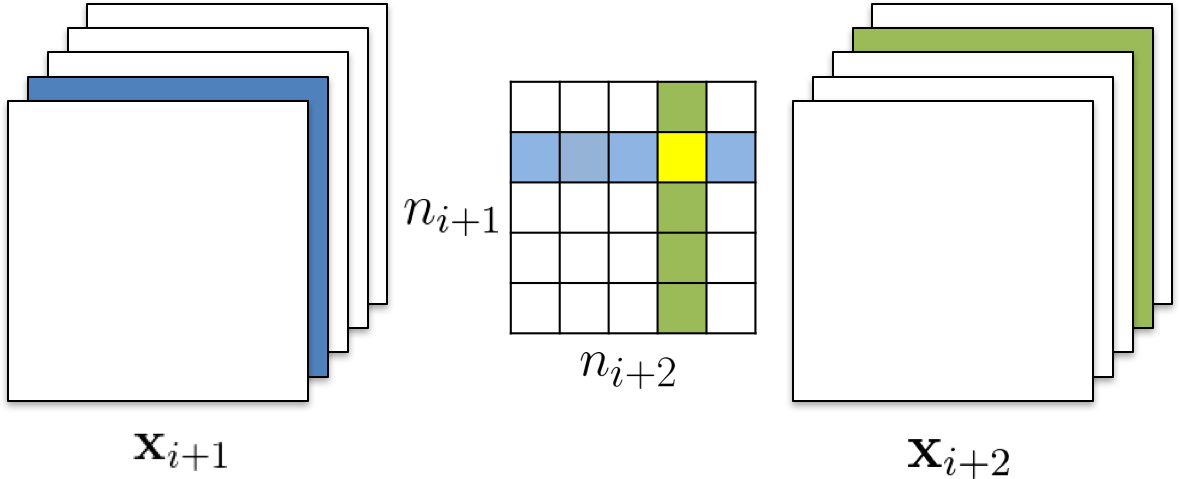}
\end{tabular}
\caption{Pruning filters across consecutive layers.
The independent pruning strategy calculates the filter sum (columns marked in green) without considering feature maps removed in previous layer (shown in blue),
so the kernel weights marked in yellow are still included.
The greedy pruning strategy does not count kernels for the already pruned feature maps.
Both approaches result in a $(n_{i+1} - 1) \times (n_{i+2} -1)$ kernel matrix.}
\label{fig:greedy_vs_independent}
\end{figure*}

Figure \ref{fig:greedy_vs_independent} illustrates the difference between two approaches in calculating the sum of absolute weights.
The greedy approach, though not globally optimal, is holistic and results in pruned networks with higher accuracy especially when many filters are pruned.

\begin{figure*}[htbp]
\centering
\begin{tabular}{l}
      \includegraphics[width=0.8\linewidth]{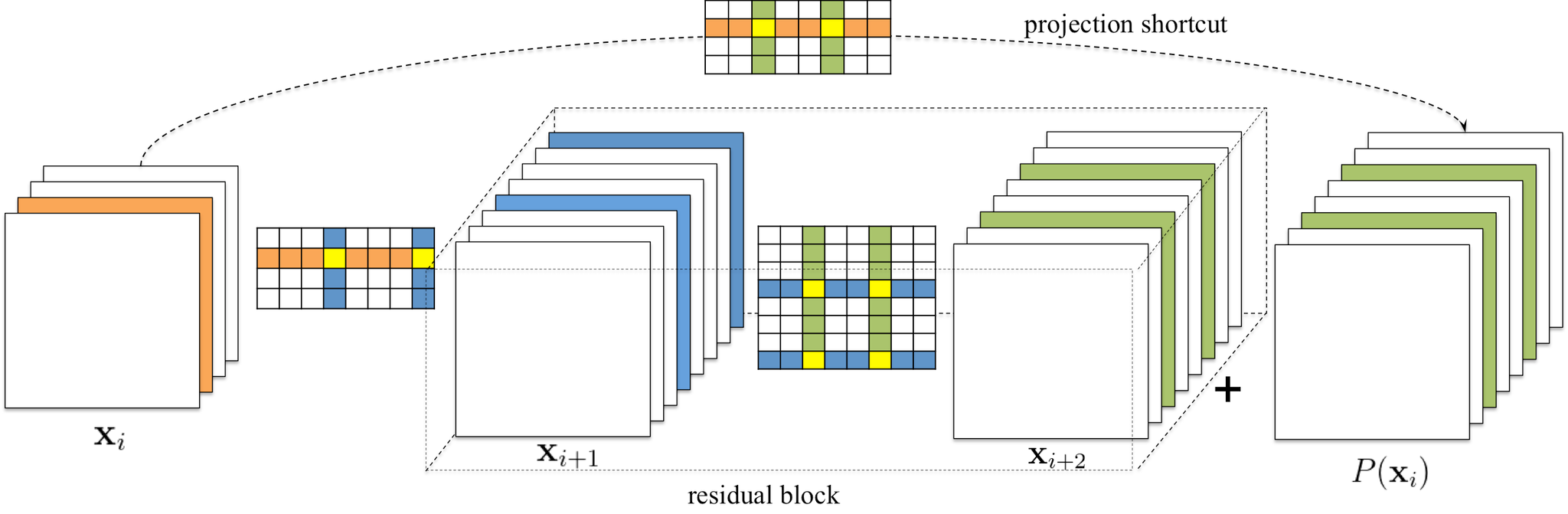}
\end{tabular}
\caption{Pruning residual blocks with the projection shortcut. 
The filters to be pruned for the second layer of the residual block (marked as green) are determined by the pruning result of the shortcut projection.
The first layer of the residual block can be pruned without restrictions.
}
\label{fig:shortcut}
\end{figure*}

For simpler CNNs like VGGNet or AlexNet, we can easily prune any of the filters in any convolutional layer.
However, for complex network architectures such as Residual networks~(\cite{resnet}), pruning filters may not be straightforward.
The architecture of ResNet imposes restrictions and the filters need to be pruned carefully.
We show the filter pruning for residual blocks with projection mapping in Figure~\ref{fig:shortcut}.
Here, the filters of the first layer in the residual block can be arbitrarily pruned, as it does not change the number of output feature maps of the block.
However, the correspondence between the output feature maps of the second convolutional layer and the identity feature maps makes it difficult to prune.
Hence, to prune the second convolutional layer of the residual block, the corresponding projected feature maps must also be pruned.
Since the identical feature maps are more important than the added residual maps, the feature maps to be pruned should be determined by the pruning results of the shortcut layer.
To determine which identity feature maps are to be pruned, we use the same selection criterion based on the filters of the shortcut convolutional layers (with $1 \times 1$ kernels).
The second layer of the residual block is pruned with the same filter index as selected by the pruning of the shortcut layer.

\subsection{Retraining pruned networks to regain accuracy}
\label{sec:retrain}

After pruning the filters, the performance degradation should be compensated by retraining the network.
There are two strategies to prune the filters across multiple layers:

1. \emph{Prune once and retrain}: Prune filters of multiple layers at once and retrain them until the original accuracy is restored.\\
2. \emph{Prune and retrain iteratively}: Prune filters layer by layer or filter by filter and then retrain iteratively.
The model is retrained before pruning the next layer for the weights to adapt to the changes from the pruning process.

We find that for the layers that are resilient to pruning, the prune and retrain once strategy can be used to prune away significant portions of the network and any loss in accuracy can be regained by retraining for a short period of time (less than the original training time). 
However, when some filters from the sensitive layers are pruned away or large portions of the networks are pruned away, it may not be possible to recover the original accuracy.
Iterative pruning and retraining may yield better results, but the iterative process requires many more epochs especially for very deep networks.
\section{Experiments}
We prune two types of networks: simple CNNs (VGG-16 on CIFAR-10) and Residual networks (ResNet-56/110 on CIFAR-10 and ResNet-34 on ImageNet).
Unlike AlexNet or VGG (on ImageNet) that are often used to demonstrate model compression, both VGG (on CIFAR-10) and Residual networks have fewer parameters in the fully connected layers. 
Hence, pruning a large percentage of parameters from these networks is challenging.
We implement our filter pruning method in Torch7~(\cite{collobert2011torch7}). 
When filters are pruned, a new model with fewer filters is created and the remaining parameters of the modified layers as well as the unaffected layers are copied into the new model.
Furthermore, if a convolutional layer is pruned, the weights of the subsequent batch normalization layer are also removed.
To get the baseline accuracies for each network, we train each model from scratch and follow the same pre-processing and hyper-parameters as ResNet~(\cite{resnet}).
For retraining, we use a constant learning rate $0.001$ and retrain 40 epochs for CIFAR-10 and 20 epochs for ImageNet, which represents one-fourth of the original training epochs.
Past work has reported up to 3$\times$ original training times to retrain pruned networks~(\cite{han2015learning}).

\begin{table}[h]
\centering
\small
\caption{Overall results. The best test/validation accuracy during the retraining process is reported.
Training a pruned model from scratch performs worse than retraining a pruned model, which may indicate the difficulty of training a network with a small capacity.}
\label{overall}
\begin{tabular}{lllllll}
\toprule
         & Model              & Error(\%)       & FLOP                &  Pruned \%  & Parameters           & Pruned \% \\ \hline
         & VGG-16             & 6.75            & $3.13 \times 10^8$  &             & $1.5 \times 10^7$    & \\
         & VGG-16-pruned-A     & \textbf{6.60}  & $2.06 \times 10^8$  &  34.2\%     & $5.4 \times 10^6$    & 64.0\% \\
         & VGG-16-pruned-A scratch-train & 6.88   &                     &             &                       &\\ \hline
         & ResNet-56             & 6.96         & $1.25 \times 10^8$  &             & $8.5 \times 10^5$     & \\
         & ResNet-56-pruned-A    & 6.90         & $1.12 \times 10^8$  &  10.4\%     & $7.7 \times 10^5$     & 9.4\% \\
         & ResNet-56-pruned-B    & \textbf{6.94}         & $9.09 \times 10^7$  &  27.6\%     & $7.3 \times 10^5$     & 13.7\% \\
         & ResNet-56-pruned-B scratch-train   & 8.69 &                  &             &                       & \\ \hline
         & ResNet-110            & 6.47         & $2.53 \times 10^8$  &             & $1.72 \times 10^6$    & \\
         & ResNet-110-pruned-A   & \textbf{6.45}         & $2.13 \times 10^8$  &  15.9\%     & $1.68 \times 10^6$    & 2.3\%  \\
         & ResNet-110-pruned-B   & 6.70         & $1.55 \times 10^8$  &  38.6\%     & $1.16 \times 10^6$    & 32.4\%  \\
         & ResNet-110-pruned-B scratch-train      & 7.06                 &             &                       & \\ \hline
         & ResNet-34             & 26.77           & $3.64 \times 10^9$  &          & $2.16 \times 10^7$    & \\
         & ResNet-34-pruned-A    & 27.44           & $3.08 \times 10^9$  & 15.5\%   & $1.99 \times 10^7$    & 7.6\%\\
         & ResNet-34-pruned-B  & 27.83           & $2.76 \times 10^9$  &  24.2\%    & $1.93 \times 10^7$    & 10.8\%\\
         & ResNet-34-pruned-C  & 27.52           & $3.37 \times 10^9$  &  7.5\%    & $2.01 \times 10^7$    & 7.2\%\\
\bottomrule
\end{tabular}
\end{table}

\subsection{VGG-16 on CIFAR-10}
VGG-16 is a high-capacity network originally designed for the ImageNet dataset~(\cite{vgg}).
Recently, \cite{torch_cifar10} applies a slightly modified version of the model on CIFAR-10 and achieves state of the art results.
As shown in Table~\ref{tab:vgg}, VGG-16 on CIFAR-10 consists of 13 convolutional layers and 2 fully connected layers, in which the fully connected layers do not occupy large portions of parameters due to the small input size and less hidden units.
We use the model described in~\cite{torch_cifar10} but add Batch Normalization~(\cite{batchnorm}) layer after each convolutional layer and the first linear layer, without using Dropout~(\cite{dropout}).
Note that when the last convolutional layer is pruned, the input to the linear layer is changed and the connections are also removed.

\begin{table}[t]
\centering
\small
\caption{VGG-16 on CIFAR-10 and the pruned model. 
The last two columns show the number of feature maps and the reduced percentage of FLOP from the pruned model.}
\label{tab:vgg}
\begin{tabular}{l|rrrr|rr}
\toprule
layer type  & $w_i\times h_i$ & \#Maps     & FLOP       & \#Params  & \#Maps  & FLOP\% \\ \hline
Conv\_1     & $32\times32$  & 64           & 1.8E+06    & 1.7E+03   & 32      & 50\%   \\
Conv\_2     & $32\times32$  & 64           & 3.8E+07    & 3.7E+04   & 64      & 50\%   \\
Conv\_3     & $16\times16$  & 128          & 1.9E+07    & 7.4E+04   & 128     & 0\%    \\
Conv\_4     & $16\times16$  & 128          & 3.8E+07    & 1.5E+05   & 128     & 0\%    \\
Conv\_5     & $8\times8$    & 256          & 1.9E+07    & 2.9E+05   & 256     & 0\%    \\
Conv\_6     & $8\times8$    & 256          & 3.8E+07    & 5.9E+05   & 256     & 0\%    \\
Conv\_7     & $8\times8$    & 256          & 3.8E+07    & 5.9E+05   & 256     & 0\%    \\
Conv\_8     & $4\times4$    & 512          & 1.9E+07    & 1.2E+06   & 256     & 50\%   \\
Conv\_9     & $4\times4$    & 512          & 3.8E+07    & 2.4E+06   & 256     & 75\%   \\
Conv\_10    & $4\times4$    & 512          & 3.8E+07    & 2.4E+06   & 256     & 75\%   \\
Conv\_11    & $2\times2$    & 512          & 9.4E+06    & 2.4E+06   & 256     & 75\%   \\
Conv\_12    & $2\times2$    & 512          & 9.4E+06    & 2.4E+06   & 256     & 75\%   \\
Conv\_13    & $2\times2$    & 512          & 9.4E+06    & 2.4E+06   & 256     & 75\%   \\
Linear      & 1             & 512          & 2.6E+05    & 2.6E+05   & 512     & 50\%   \\
Linear      & 1             & 10           & 5.1E+03    & 5.1E+03   & 10      & 0\%    \\ \hline
Total       &               &              & 3.1E+08    & 1.5E+07   &         & 34\%   \\
\bottomrule
\end{tabular}
\vspace{-2mm}
\end{table}

\begin{figure*}[h]
\centering
\begin{tabular}{l}
 \includegraphics[width=0.8\linewidth]{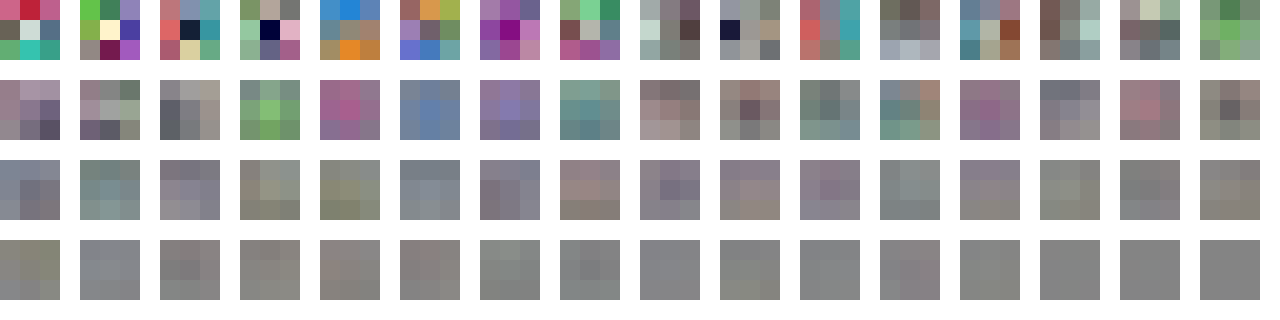}
\end{tabular}
\caption{Visualization of filters in the first convolutional layer of VGG-16 trained on CIFAR-10. Filters are ranked by $\ell_1$-norm.}
\label{fig:visual_filters}
\end{figure*}

As shown in Figure~\ref{fig:prune_smallest_filter}, each of the convolutional layers with 512 feature maps can drop at least 60\% of filters without affecting the accuracy.
Figure~\ref{fig:retrain_vgg} shows that with retraining, almost 90\% of the filters of these layers can be safely removed.
One possible explanation is that these filters operate on $4\times 4$ or $2\times 2$ feature maps, which may have no meaningful spatial connections in such small dimensions.
For instance, ResNets for CIFAR-10 do not perform any convolutions for feature maps below $8 \times 8$ dimensions.
Unlike previous work~(\cite{zeiler2014visualizing, han2015learning}), we observe that the first layer is robust to pruning as compared to the next few layers.
This is possible for a simple dataset like CIFAR-10, on which the model does not learn as much useful filters as on ImageNet (as shown in Figure.~\ref{fig:visual_filters}).
Even when 80\% of the filters from the first layer are pruned, the number of remaining filters (12) is still larger than the number of raw input channels.
However, when removing 80\% filters from the second layer, the layer corresponds to a 64 to 12 mapping, which may lose significant information from previous layers, thereby hurting the accuracy.
With 50\% of the filters being pruned in layer 1 and from 8 to 13, we achieve 34\% FLOP reduction for the same accuracy.

\subsection{ResNet-56/110 on CIFAR-10}
\begin{figure*}[tbp]
\centering
\begin{tabular}{l}
 \includegraphics[width=0.33\linewidth]{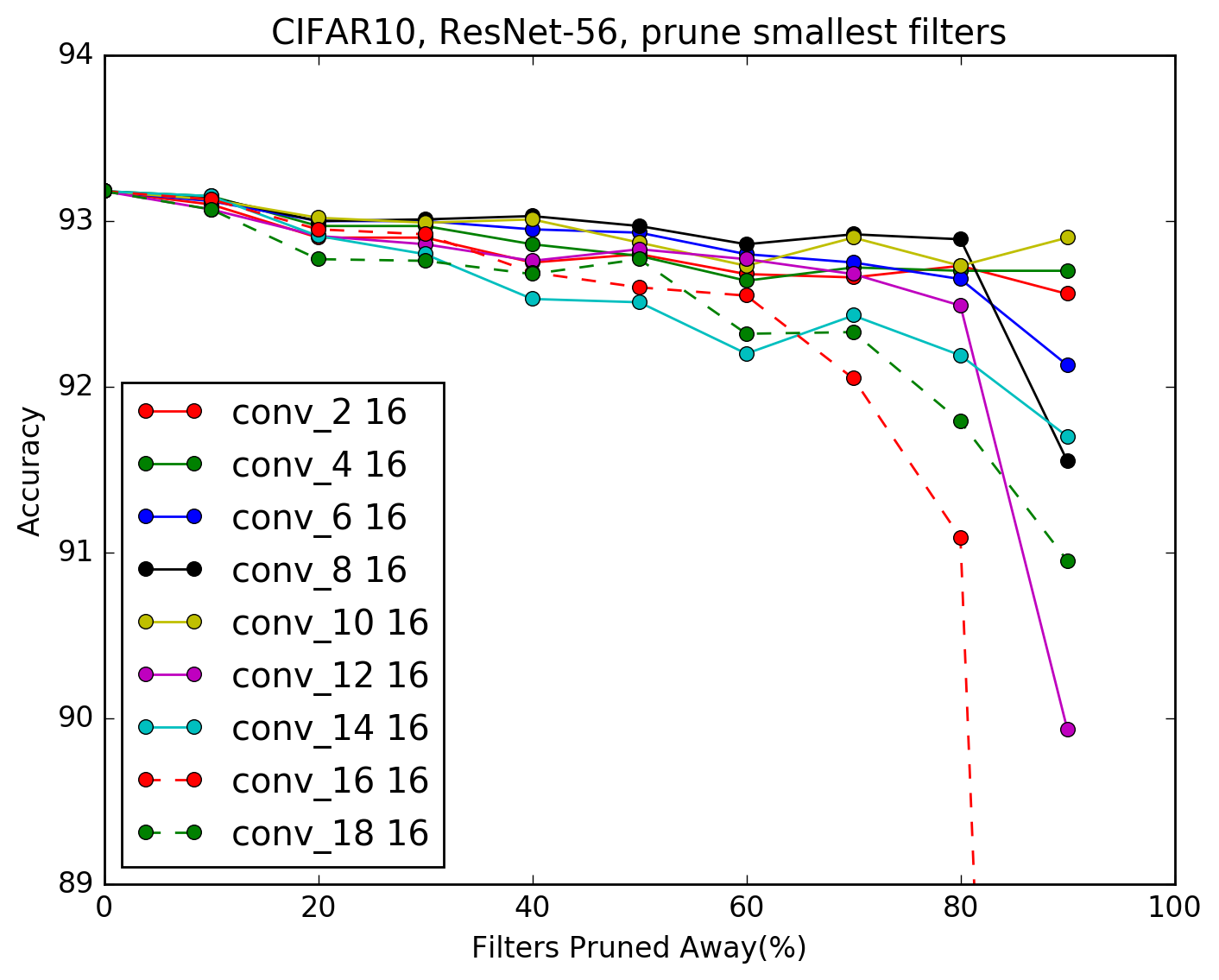}
 \includegraphics[width=0.33\linewidth]{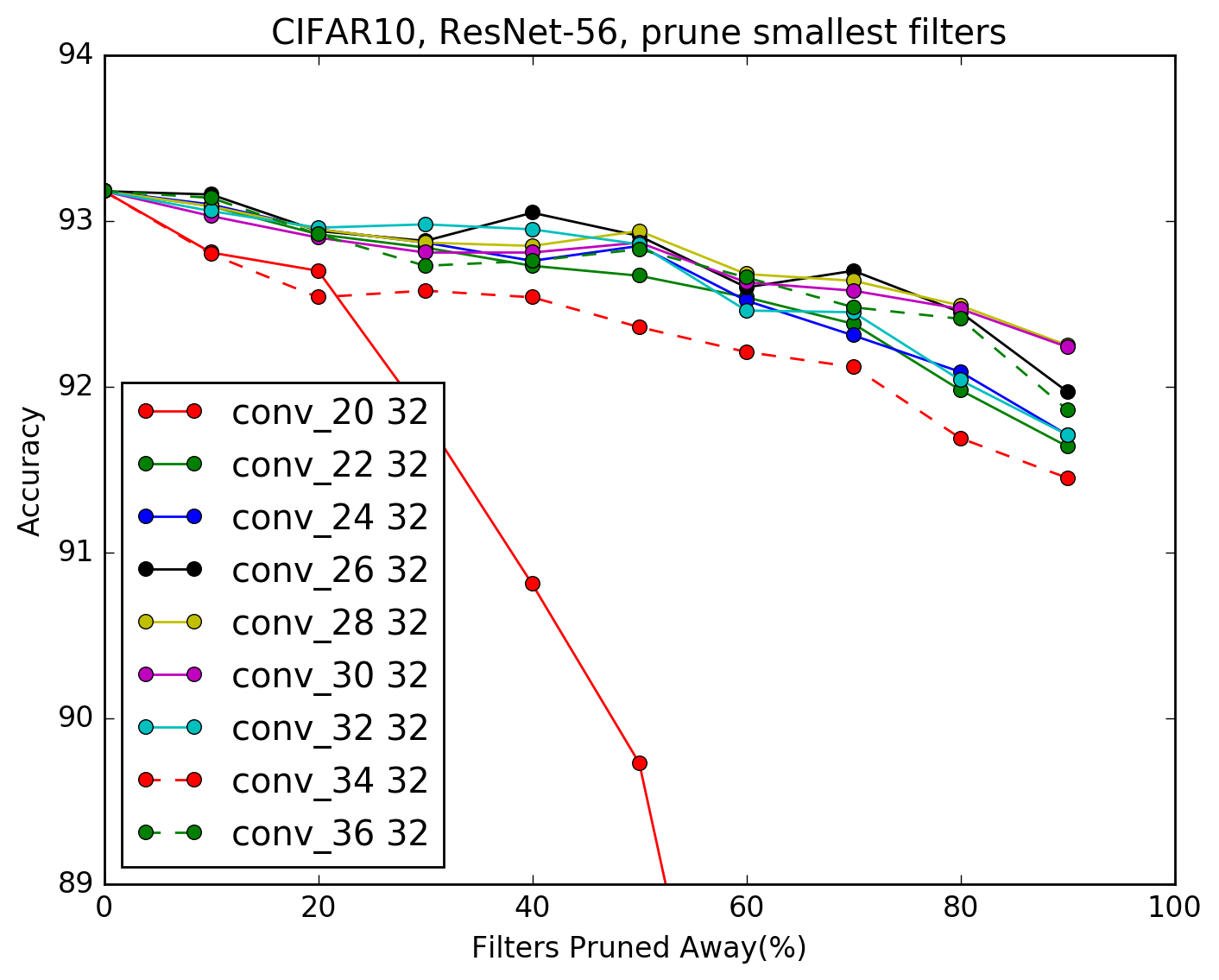} 
 \includegraphics[width=0.33\linewidth]{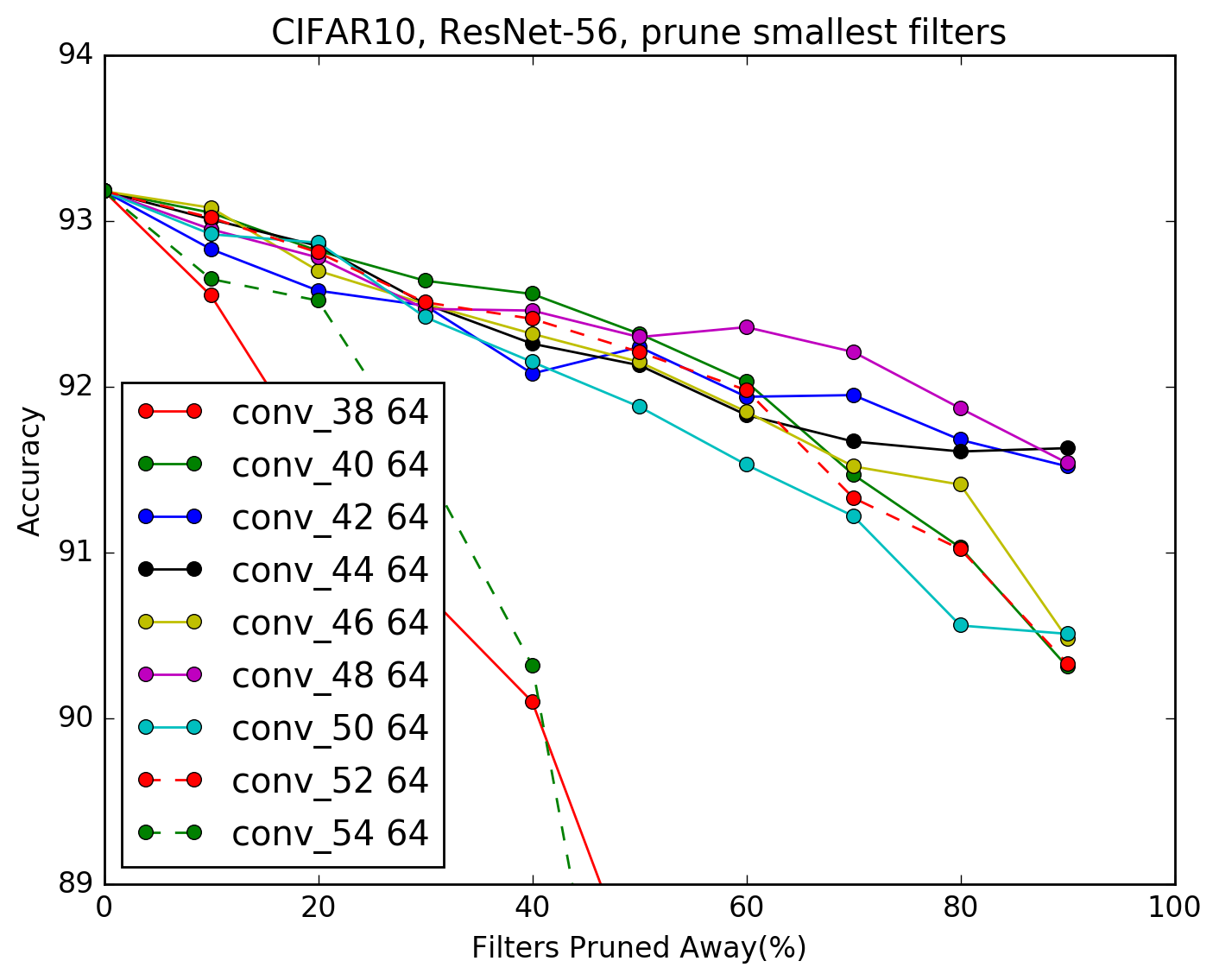}\\
 \includegraphics[width=0.33\linewidth]{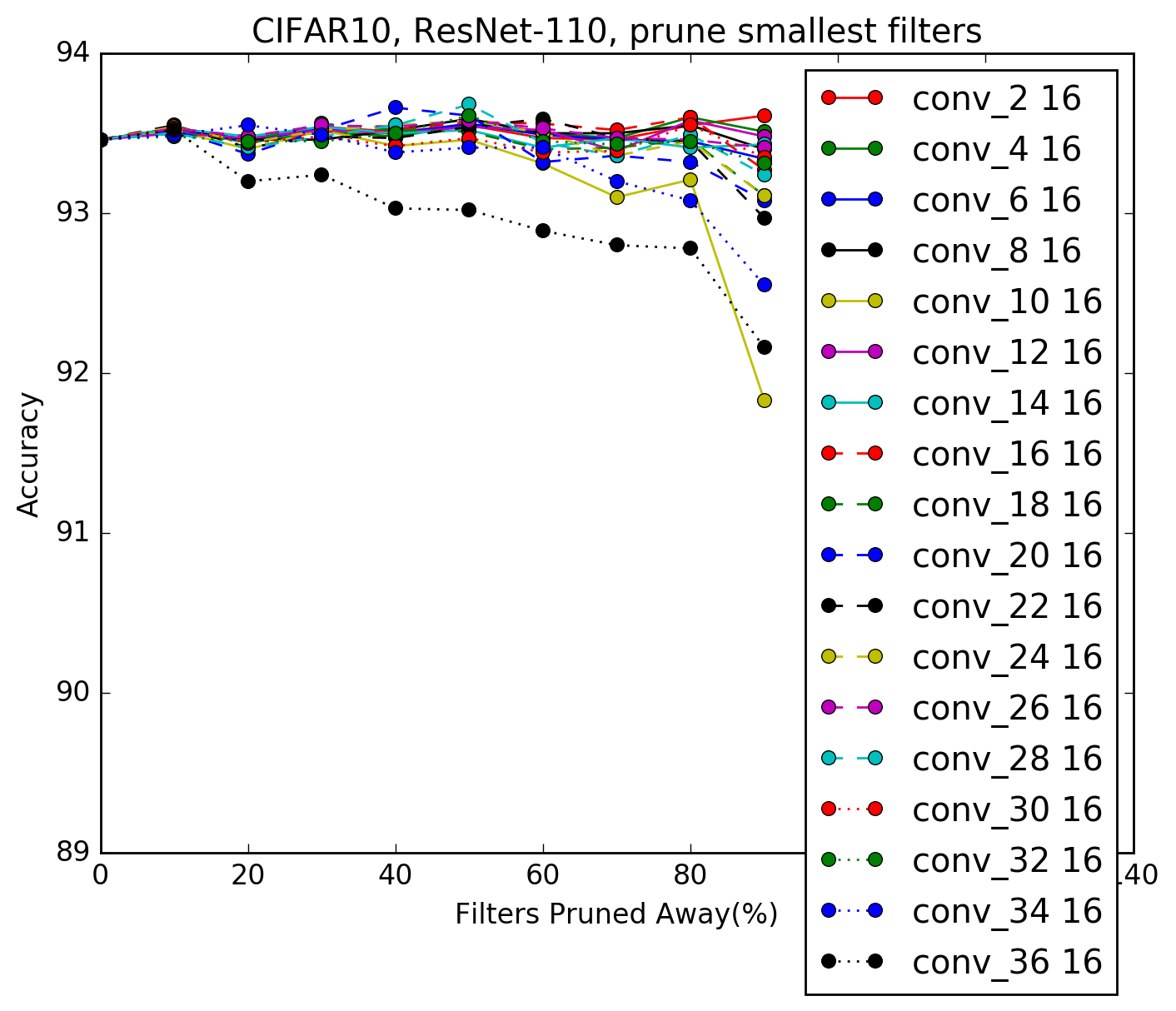}
 \includegraphics[width=0.33\linewidth]{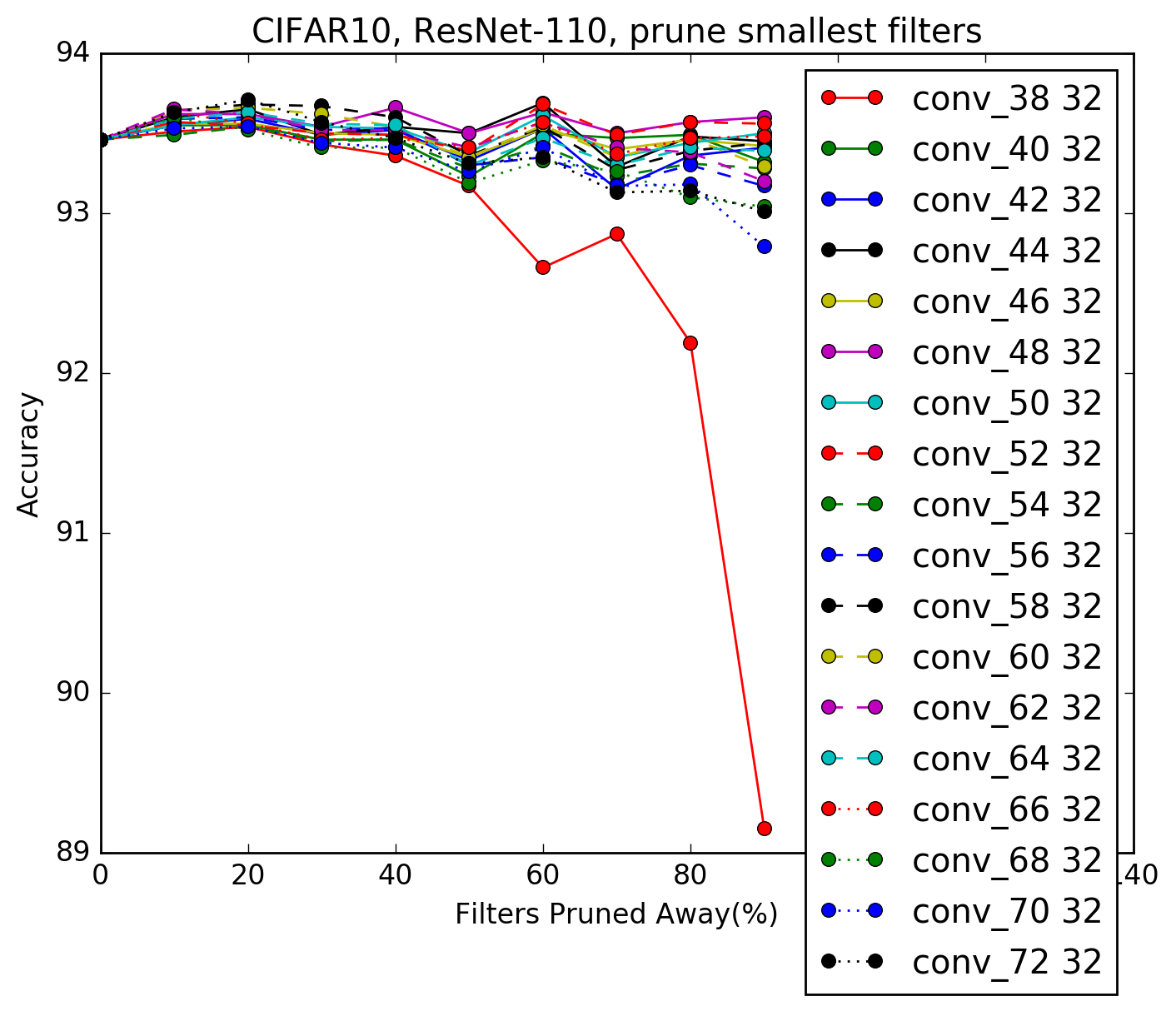}
 \includegraphics[width=0.33\linewidth]{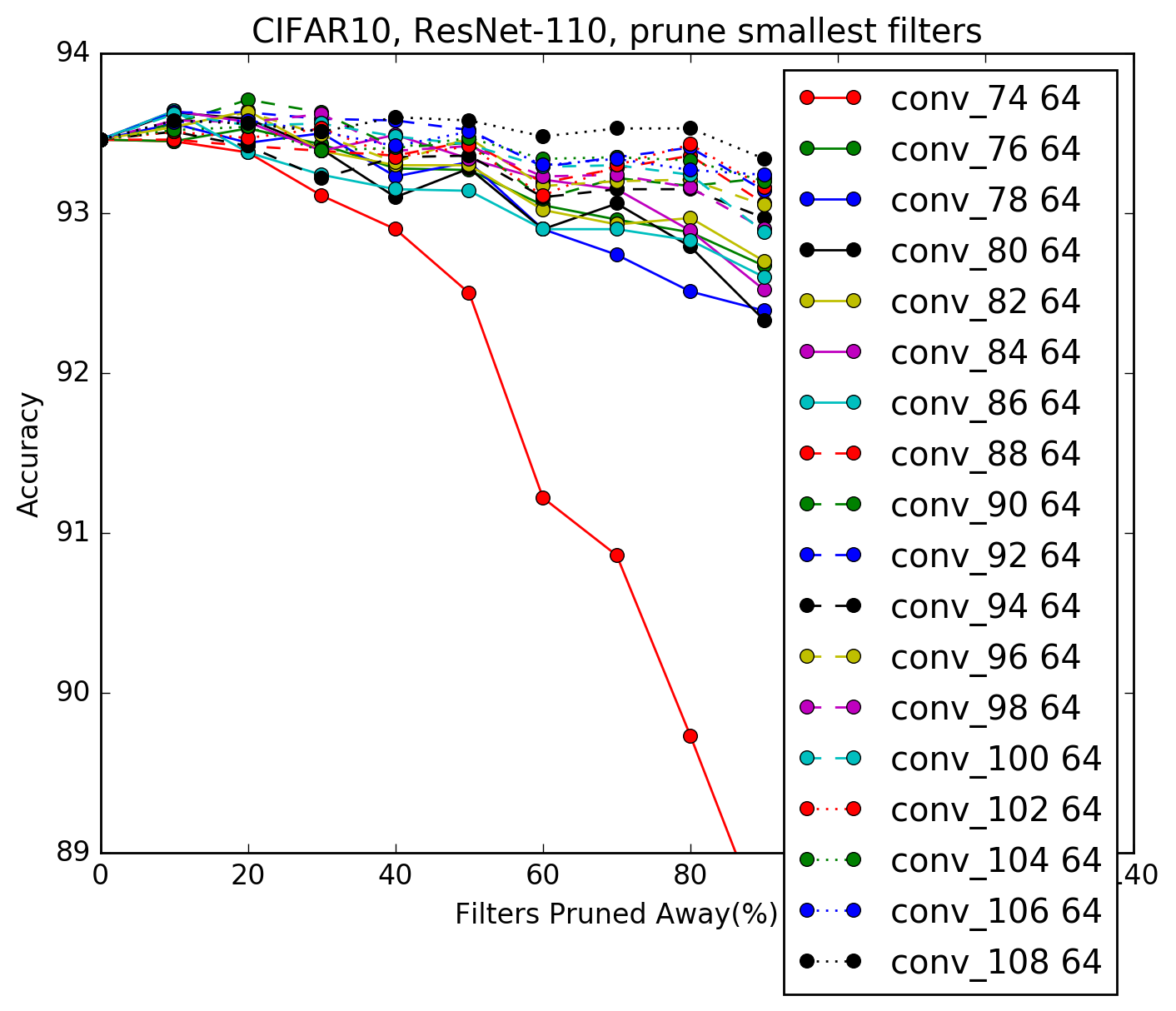}
\end{tabular}
\caption{Sensitivity to pruning for the first layer of each residual block of ResNet-56/110.}
\label{fig:prune_resnet-56/110}
\end{figure*}

ResNets for CIFAR-10 have three stages of residual blocks for feature maps with sizes of $32 \times 32$, $16 \times 16$ and $8 \times 8$.
Each stage has the same number of residual blocks.
When the number of feature maps increases, the shortcut layer provides an identity mapping with an additional zero padding for the increased dimensions.
Since there is no projection mapping for choosing the identity feature maps, we only consider pruning the first layer of the residual block.
As shown in Figure~\ref{fig:prune_resnet-56/110}, most of the layers are robust to pruning.
For ResNet-110, pruning some single layers without retraining even improves the performance.
In addition, we find that layers that are sensitive to pruning (layers 20, 38 and 54 for ResNet-56, layer 36, 38 and 74 for ResNet-110) lie at the residual blocks close to the layers where the number of feature maps changes, e.g., the first and the last residual blocks for each stage.
We believe this happens because the precise residual errors are necessary for the newly added empty feature maps.

The retraining performance can be improved by skipping these sensitive layers.
As shown in Table~\ref{overall}, ResNet-56-pruned-A improves the performance by pruning 10\% filters while skipping the sensitive layers 16, 20, 38 and 54.
In addition, we find that deeper layers are more sensitive to pruning than layers in the earlier stages of the network.
Hence, we use a different pruning rate for each stage.
We use $p_i$ to denote the pruning rate for layers in the $i$th stage.
ResNet-56-pruned-B skips more layers (16, 18, 20, 34, 38, 54) and prunes layers with $p_1$=60\%, $p_2$=30\% and $p_3$=10\%.
For ResNet-110, the first pruned model gets a slightly better result with $p_1$=50\% and layer 36 skipped.
ResNet-110-pruned-B skips layers 36, 38, 74 and prunes with $p_1$=50\%, $p_2$=40\% and $p_3$=30\%.
When there are more than two residual blocks at each stage, the middle residual blocks may be redundant and can be easily pruned.
This might explain why ResNet-110 is easier to prune than ResNet-56.

\subsection{ResNet-34 on ILSVRC2012}
\begin{figure*}[t]
\centering
\begin{tabular}{l}
\subfigure[Pruning the first layer of residual blocks]{
      \includegraphics[width=0.45\linewidth]{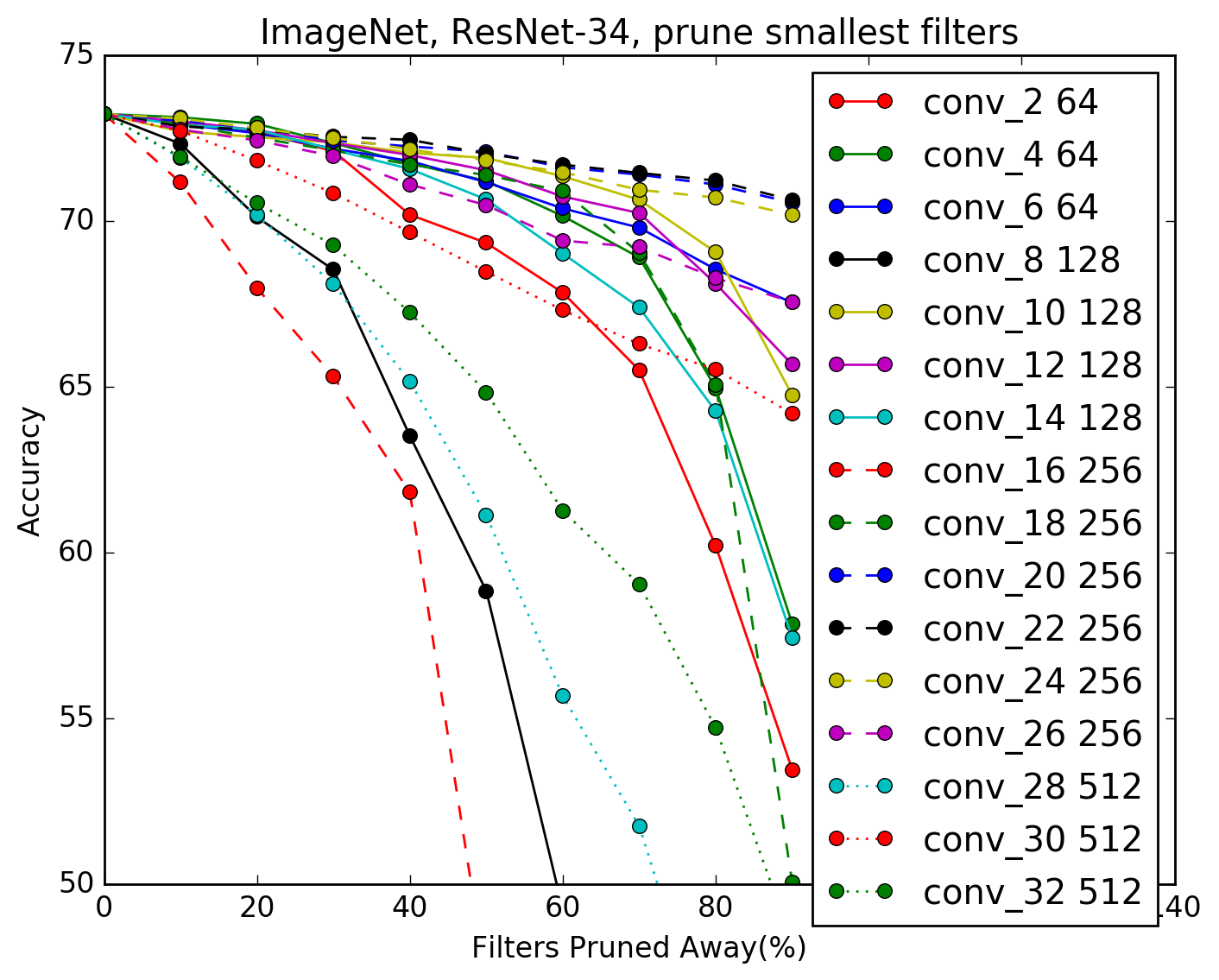}
}
\subfigure[Pruning the second layer of residual blocks]{
      \includegraphics[width=0.45\linewidth]{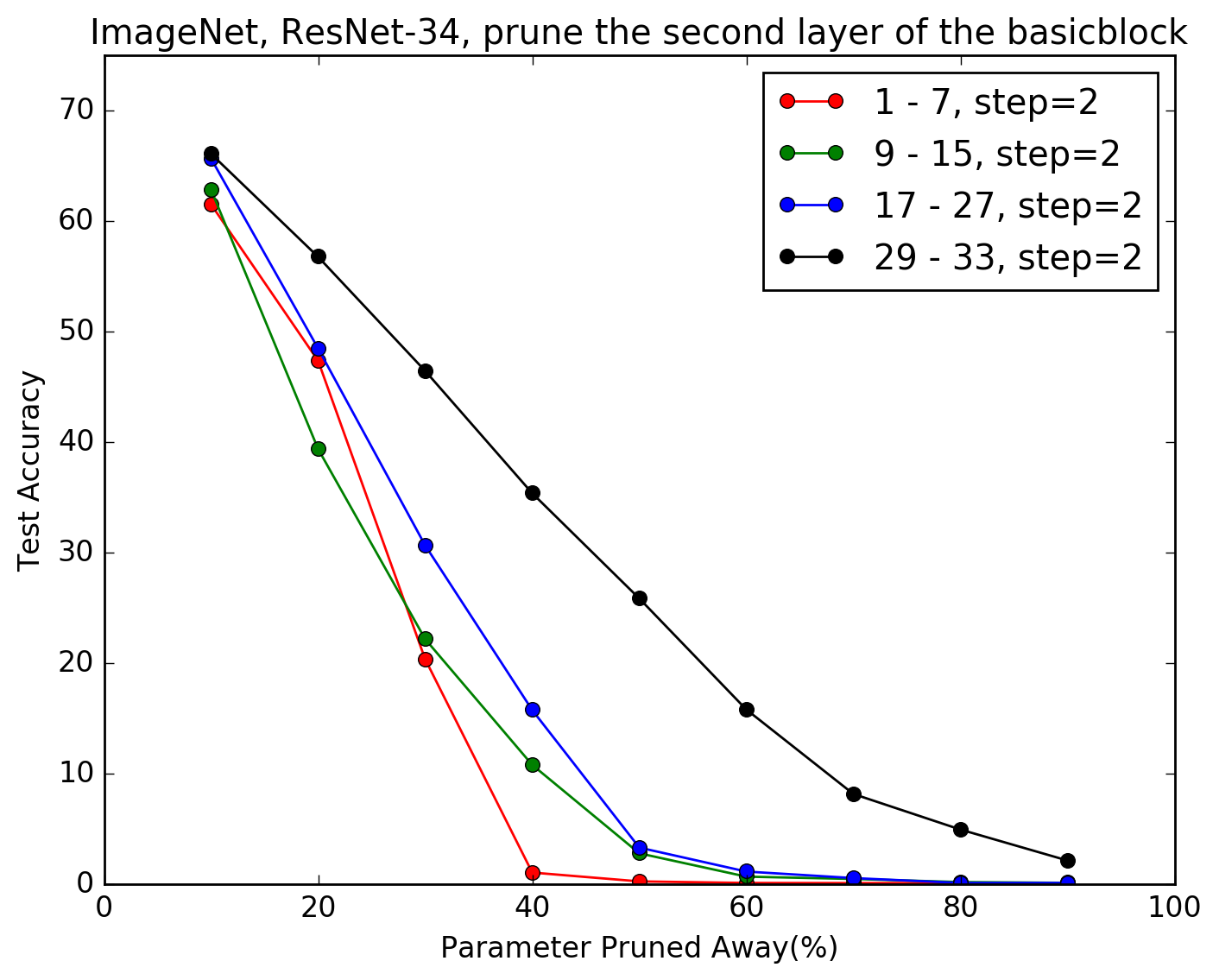}
      \label{fig:prune_resnet34:b}
}
\end{tabular}
\caption{Sensitivity to pruning for the residual blocks of ResNet-34.}
\label{fig:prune_resnet34}
\vspace{-3mm}
\end{figure*}

ResNets for ImageNet have four stages of residual blocks for feature maps with sizes of $56 \times 56$, $28 \times 28$, $14 \times 14$ and $7 \times 7$.
ResNet-34 uses the projection shortcut when the feature maps are down-sampled.
We first prune the first layer of each residual block.
Figure~\ref{fig:prune_resnet34} shows the sensitivity of the first layer of each residual block.
Similar to ResNet-56/110, the first and the last residual blocks of each stage are more sensitive to pruning than the intermediate blocks (i.e., layers 2, 8, 14, 16, 26, 28, 30, 32).
We skip those layers and prune the remaining layers at each stage equally.
In Table~\ref{overall} we compare two configurations of pruning percentages for the first three stages:
(A) $p_1$=30\%, $p_2$=30\%, $p_3$=30\%; (B) $p_1$=50\%, $p_2$=60\%, $p_3$=40\%.
Option-B provides 24\% FLOP reduction with about 1\% loss in accuracy.
As seen in the pruning results for ResNet-50/110, we can predict that ResNet-34 is relatively more difficult to prune as compared to deeper ResNets.

We also prune the identity shortcuts and the second convolutional layer of the residual blocks.
As these layers have the same number of filters, they are pruned equally.
As shown in Figure~\ref{fig:prune_resnet34:b}, these layers are more sensitive to pruning than the first layers.
With retraining, ResNet-34-pruned-C prunes the third stage with $p_3$=20\% and results in 7.5\% FLOP reduction with 0.75\% loss in accuracy.
Therefore, pruning the first layer of the residual block is more effective at reducing the overall FLOP than pruning the second layer.
This finding also correlates with the bottleneck block design for deeper ResNets,
which first reduces the dimension of input feature maps for the residual layer and then increases the dimension to match the identity mapping.

\subsection{Comparison with pruning random filters and largest filters}
\label{sec:random_largest_filters}
\begin{figure*}[h]
\centering
\begin{tabular}{l}
      \includegraphics[width=0.33\linewidth]{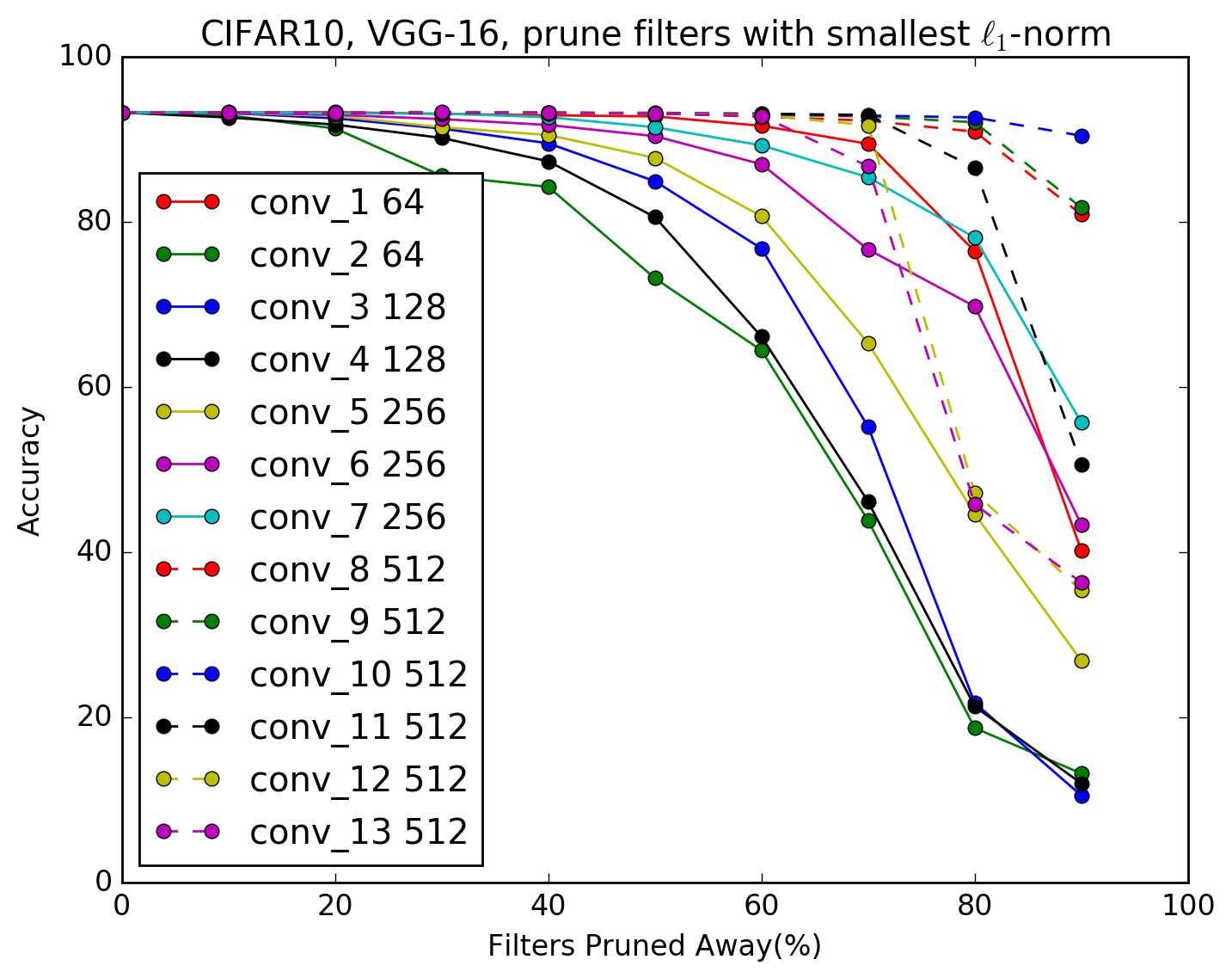}
      \includegraphics[width=0.33\linewidth]{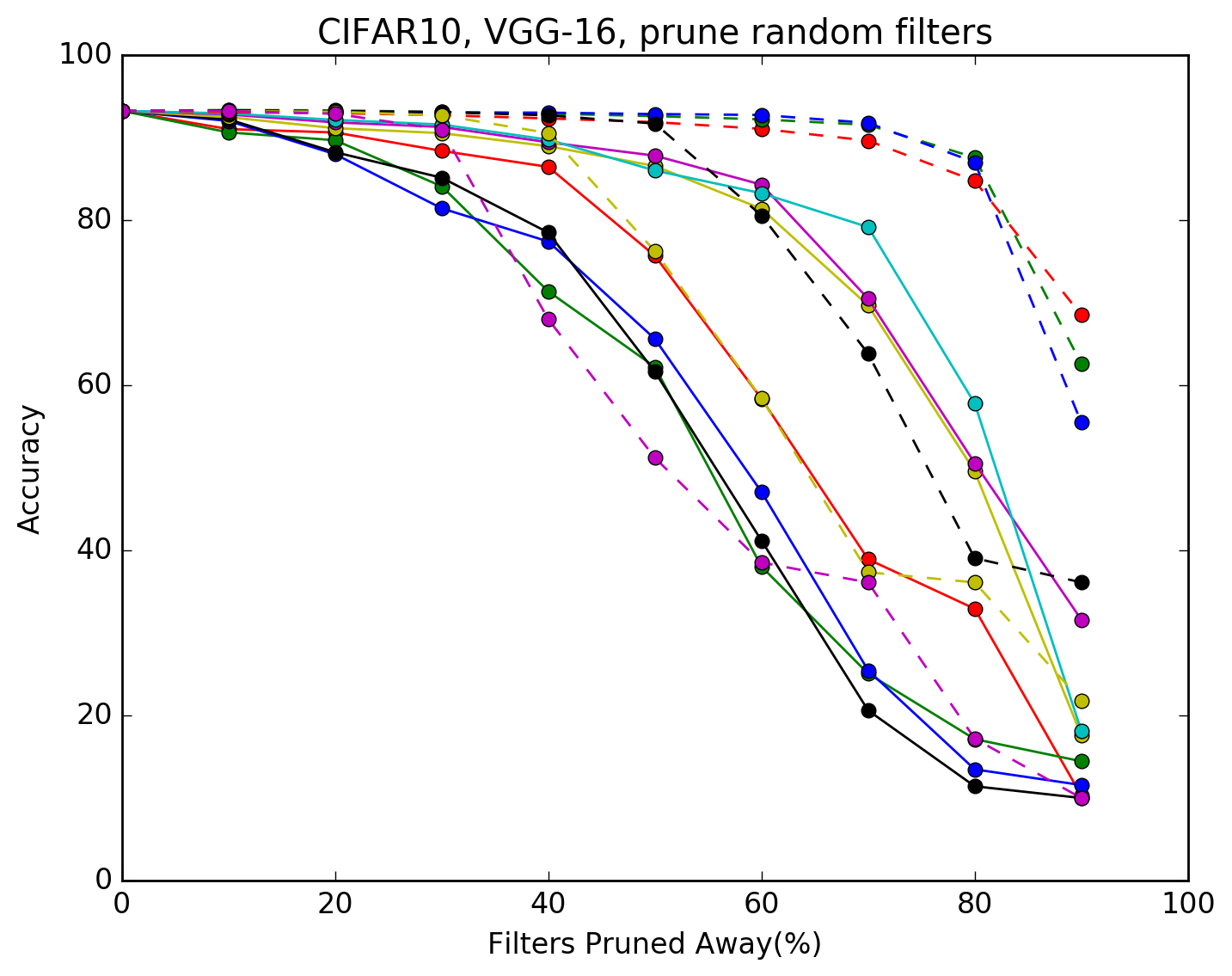}
      \includegraphics[width=0.33\linewidth]{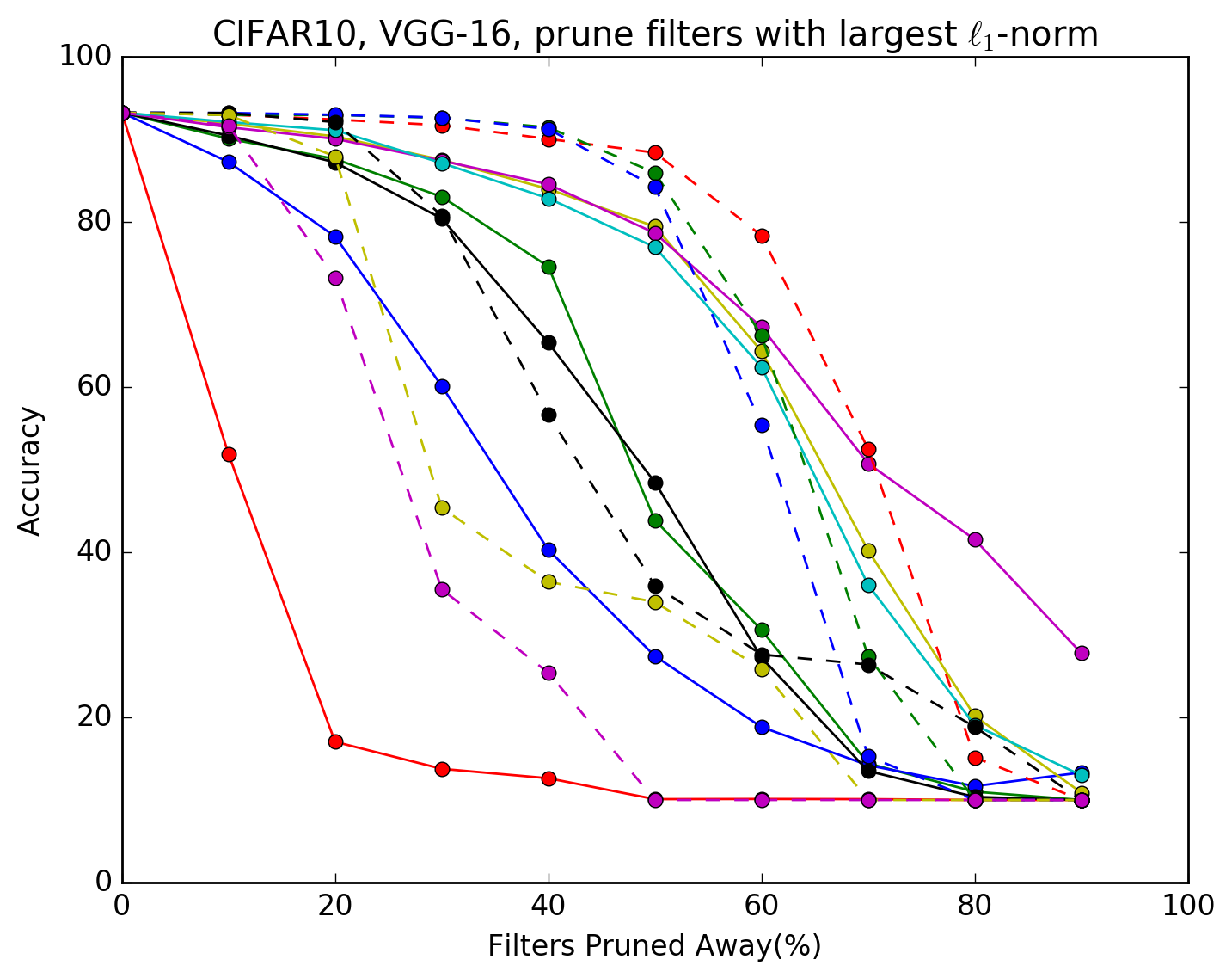}\\
\end{tabular}
\caption{Comparison of three pruning methods for VGG-16 on CIFAR-10: pruning the smallest filters, pruning random filters and pruning the largest filters. 
In random filter pruning, the order of filters to be pruned is randomly permuted.
}
\label{fig:prune_order}
\end{figure*}

We compare our approach with pruning random filters and largest filters.
As shown in Figure~\ref{fig:prune_order}, pruning the smallest filters outperforms pruning random filters for most of the layers at different pruning ratios. 
For example, smallest filter pruning has better accuracy than random filter pruning for all layers with the pruning ratio of 90\%.
The accuracy of pruning filters with the largest $\ell_1$-norms drops quickly as the pruning ratio increases, which indicates the importance of filters with larger $\ell_1$-norms.

\subsection{Comparison with activation-based feature map pruning}
\label{sec:pruning_activation}
The activation-based feature map pruning method removes the feature maps with weak activation patterns and their corresponding filters and kernels~(\cite{polyak2015channel}), which needs sample data as input to determine which feature maps to prune.
A feature map $\mathbf{x}_{i+1,j} \in \mathbb{R}^{w_{i+1}\times h_{i+1}}$ is generated by applying filter $\mathcal{F}_{i,j}\in \mathbb{R}^{n_i \times k\times k}$ to feature maps of previous layer $\mathbf{x}_i\in \mathbb{R}^{n_i \times w_i\times h_i}$, i.e., $\mathbf{x}_{i+1,j} = \mathcal{F}_{i,j}*\mathbf{x}_i$. 
Given $N$ randomly selected images $\{\mathbf{x}_1^n\}_{n=1}^N$ from the training set, the statistics of each feature map can be estimated with one epoch forward pass of the $N$ sampled data. 
Note that we calculate statistics on the feature maps generated from the convolution operations before batch normalization or non-linear activation.
We compare our $\ell_1$-norm based filter pruning with feature map pruning using the following criteria:
$\sigma_{\texttt{mean-mean}}(\mathbf{x}_{i,j}) = \frac{1}{N}\sum_{n=1}^{N}\texttt{mean}(\mathbf{x}_{i,j}^{n})$, 
$\sigma_{\texttt{mean-std}}(\mathbf{x}_{i,j}) = \frac{1}{N}\sum_{n=1}^{N}\texttt{std}(\mathbf{x}_{i,j}^{n})$,
$\sigma_{\texttt{mean-}\ell_1}(\mathbf{x}_{i,j}) = \frac{1}{N}\sum_{n=1}^{N}\|\mathbf{x}_{i,j}^{n}\|_1$,
$\sigma_{\texttt{mean-}\ell_2}(\mathbf{x}_{i,j}) = \frac{1}{N}\sum_{n=1}^{N}\|\mathbf{x}_{i,j}^{n}\|_2$ and 
$\sigma_{\texttt{var-}\ell_2}(\mathbf{x}_{i,j}) = \texttt{var}(\{\|\mathbf{x}_{i,j}^{n}\|_2\}_{n=1}^{N})$,
where \texttt{mean}, \texttt{std} and \texttt{var} are standard statistics (average, standard deviation and variance) of the input.
Here, $\sigma_{\texttt{var-}\ell_2}$ is the \emph{contribution variance of channel} criterion proposed in~\cite{polyak2015channel}, 
which is motivated by the intuition that an unimportant feature map has almost similar outputs for the whole training data and acts like an additional bias.

\begin{figure*}[tbp]
\centering
\begin{tabular}{l}
   \subfigure[$\|\mathcal{F}_{i,j}\|_1$]{
      \includegraphics[width=0.33\linewidth]{figure/vgg_bn,prune_test_single_layer_filter_l1_smallest.png}
   }
   \subfigure[$\sigma_{\texttt{mean-mean}}$]{
      \includegraphics[width=0.33\linewidth]{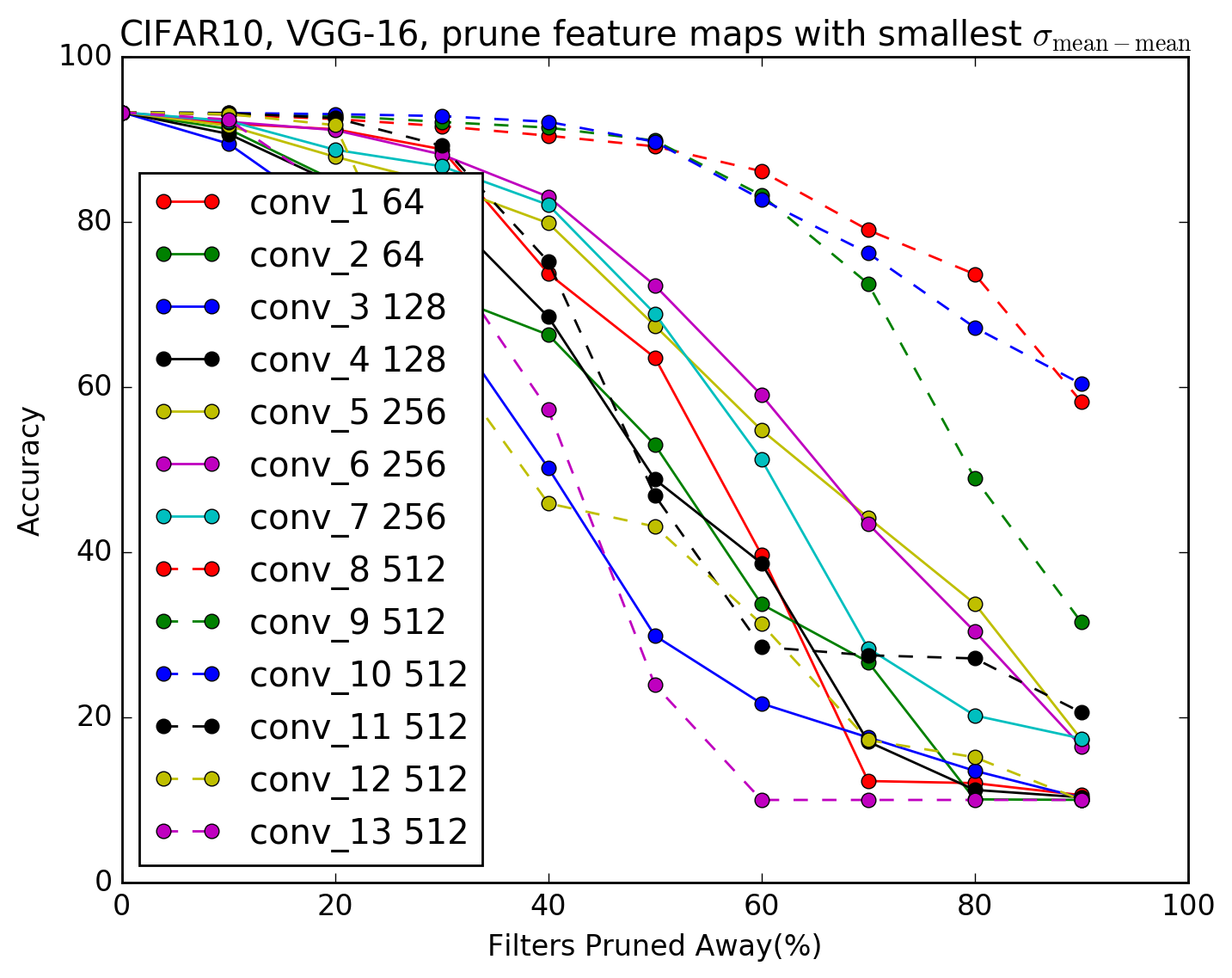}
   }
   \subfigure[$\sigma_{\texttt{mean-std}}$]{
      \includegraphics[width=0.33\linewidth]{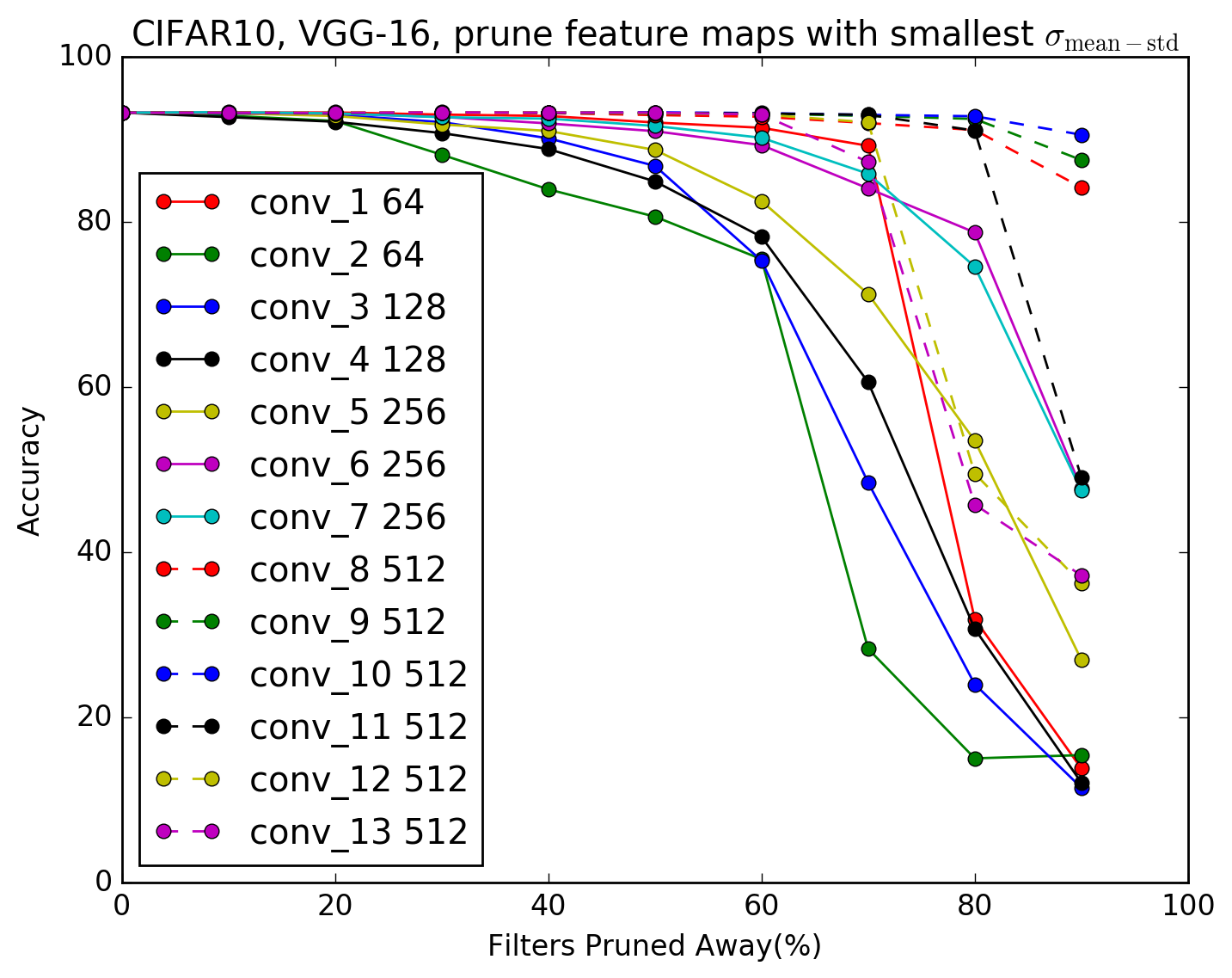}
   }\\
   \subfigure[$\sigma_{\texttt{mean-}\ell_1}$]{
      \includegraphics[width=0.33\linewidth]{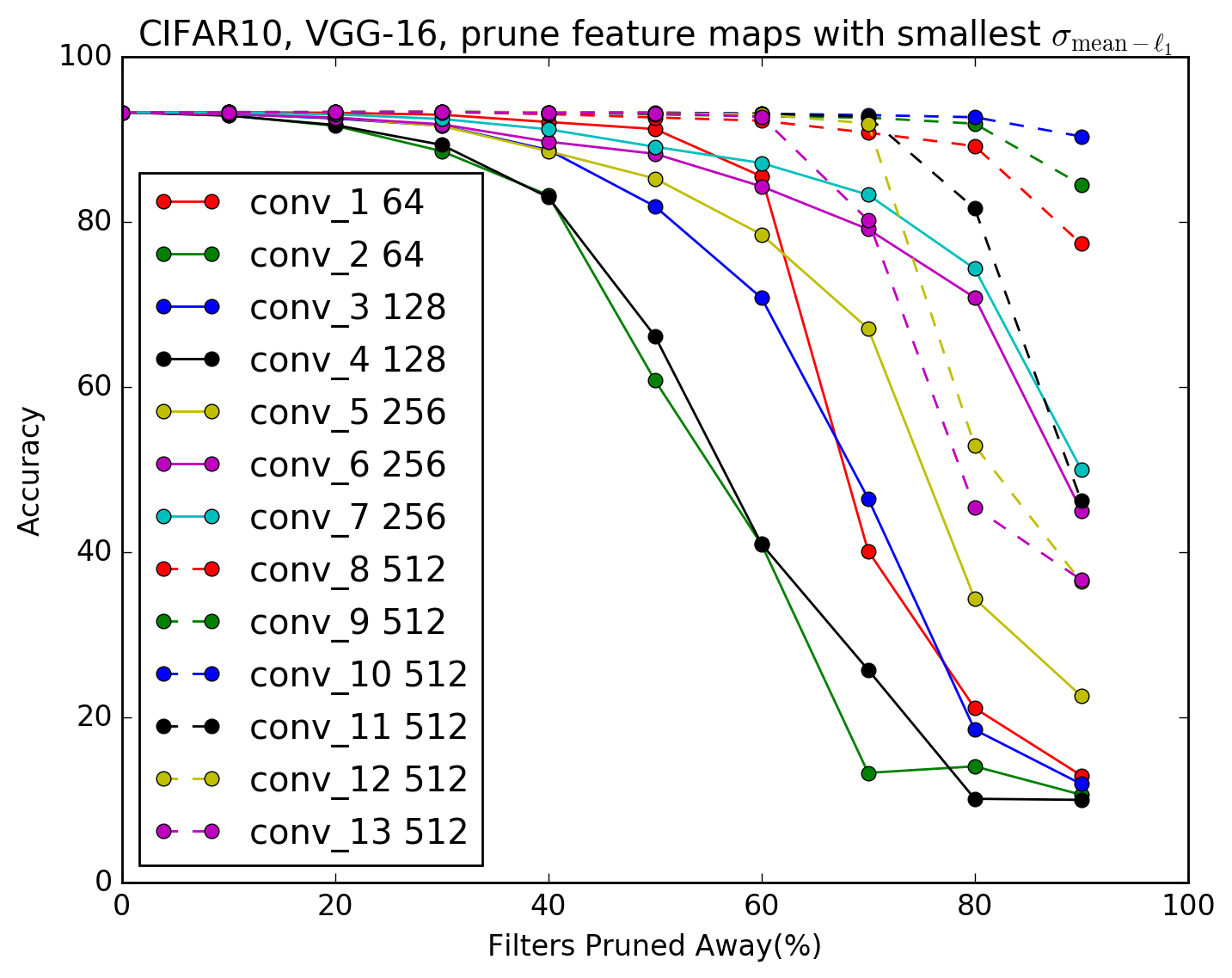}
   }
   \subfigure[$\sigma_{\texttt{mean-}\ell_2}$]{
      \includegraphics[width=0.33\linewidth]{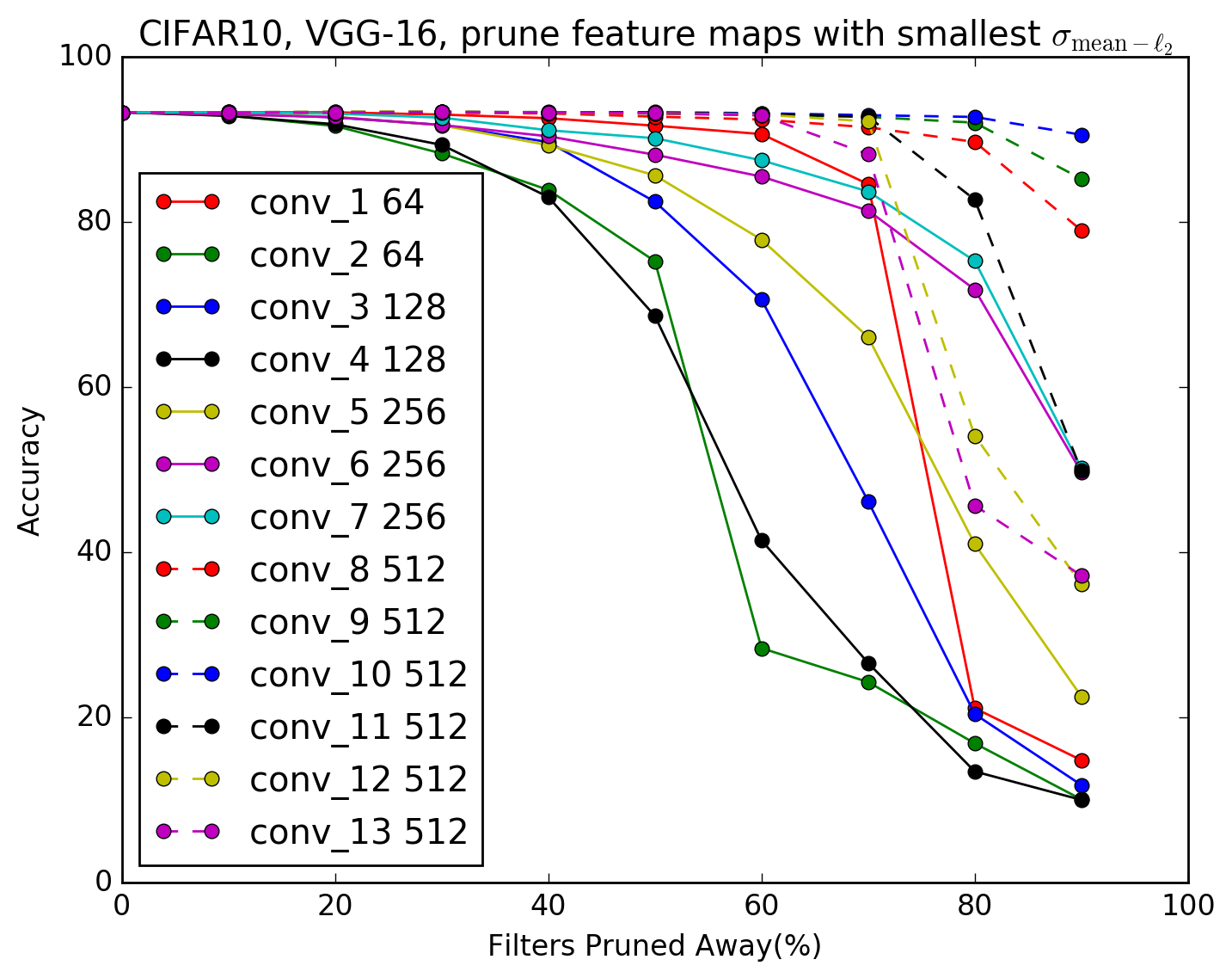}
   }
   \subfigure[$\sigma_{\texttt{var-}\ell_2}$]{
      \includegraphics[width=0.33\linewidth]{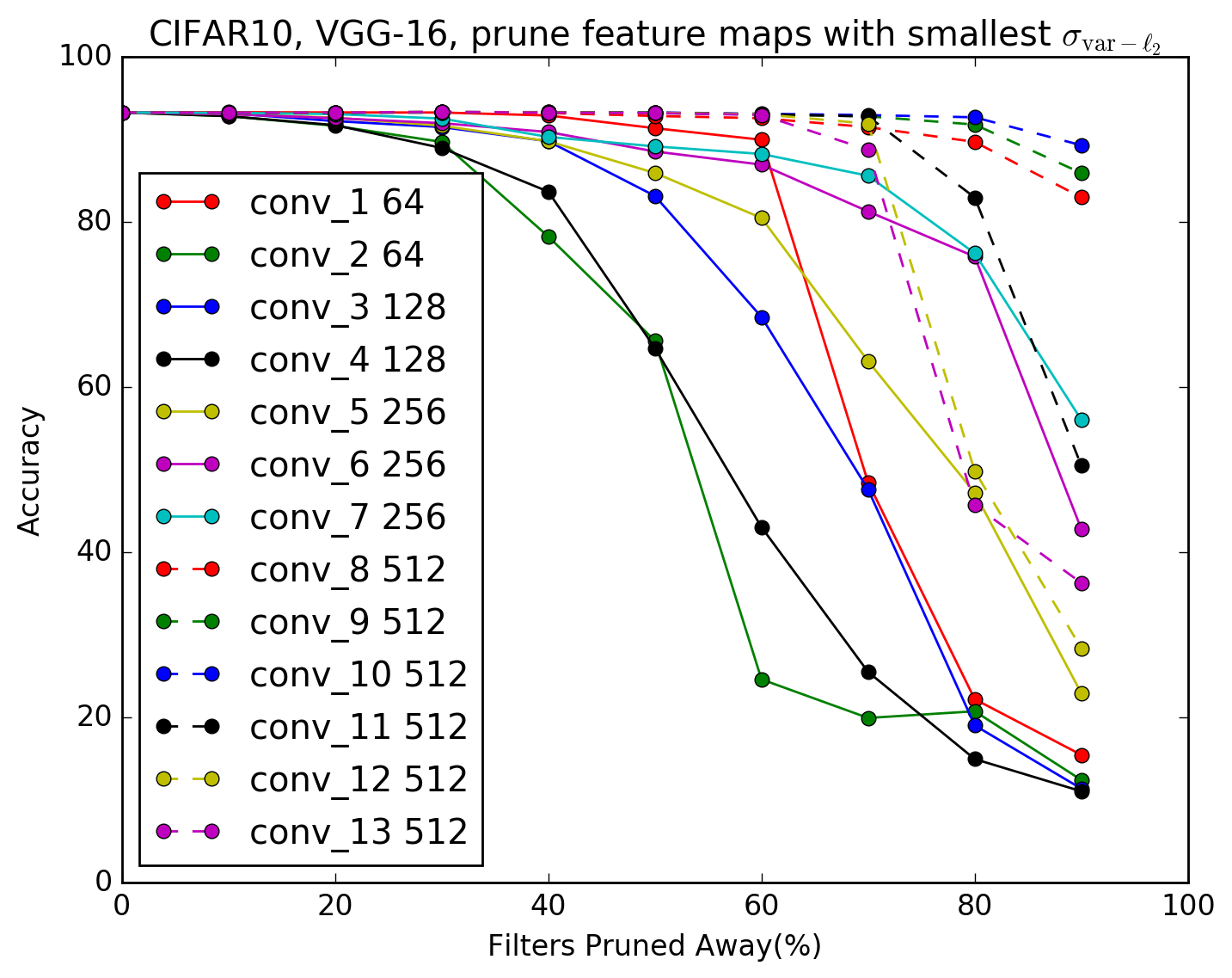}
   }
\end{tabular}
\caption{Comparison of activation-based feature map pruning for VGG-16 on CIFAR-10.}
\label{fig:prune_activation}
\end{figure*}

The estimation of the criteria becomes more accurate when more sample data is used.
Here we use the whole training set ($N=50,000$ for CIFAR-10) to compute the statistics.
The performance of feature map pruning with above criteria for each layer is shown in Figure~\ref{fig:prune_activation}. 
   Smallest filter pruning outperforms feature map pruning with the criteria $\sigma_{\texttt{mean-mean}}$, $\sigma_{\texttt{mean-}\ell_1}$, $\sigma_{\texttt{mean-}\ell_2}$ and $\sigma_{\texttt{var-}\ell_2}$.   
   The $\sigma_{\texttt{mean-std}}$ criterion has better or similar performance to $\ell_1$-norm up to pruning ratio of 60\%.
   However, its performance drops quickly after that especially for layers of $\texttt{conv\_1}$, $\texttt{conv\_2}$ and $\texttt{conv\_3}$.
We find $\ell_1$-norm is a good heuristic for filter selection considering that it is data free.

\section{Conclusions}
\label{sec:disconc}
Modern CNNs often have high capacity with large training and inference costs. 
In this paper we present a method to prune filters with relatively low weight magnitudes to produce CNNs with reduced computation costs without introducing irregular sparsity. 
It achieves about 30\% reduction in FLOP for VGGNet (on CIFAR-10) and deep ResNets without significant loss in the original accuracy.
Instead of pruning with specific layer-wise hayperparameters and time-consuming iterative retraining, we use the one-shot pruning and retraining strategy for simplicity and ease of implementation.
By performing lesion studies on very deep CNNs, we identify layers that are robust or sensitive to pruning, which can be useful for further understanding and improving the architectures.

\section*{Acknowledgments}
The authors would like to thank the anonymous reviewers for their valuable feedback.

\bibliography{iclr2017_conference}
\bibliographystyle{iclr2017_conference}

\newpage
\section{Appendix}

\subsection{Comparison with $\ell_2$-norm based filter pruning}
\label{sec:pruning_filters_l2}
\begin{figure*}[hbp]
\centering
\begin{tabular}{l}
   \subfigure[$\|\mathcal{F}_{i,j}\|_1$]{
      \includegraphics[width=0.4\linewidth]{figure/vgg_bn,prune_test_single_layer_filter_l1_smallest.png}
   }
   \subfigure[$\|\mathcal{F}_{i,j}\|_2$]{
      \includegraphics[width=0.4\linewidth]{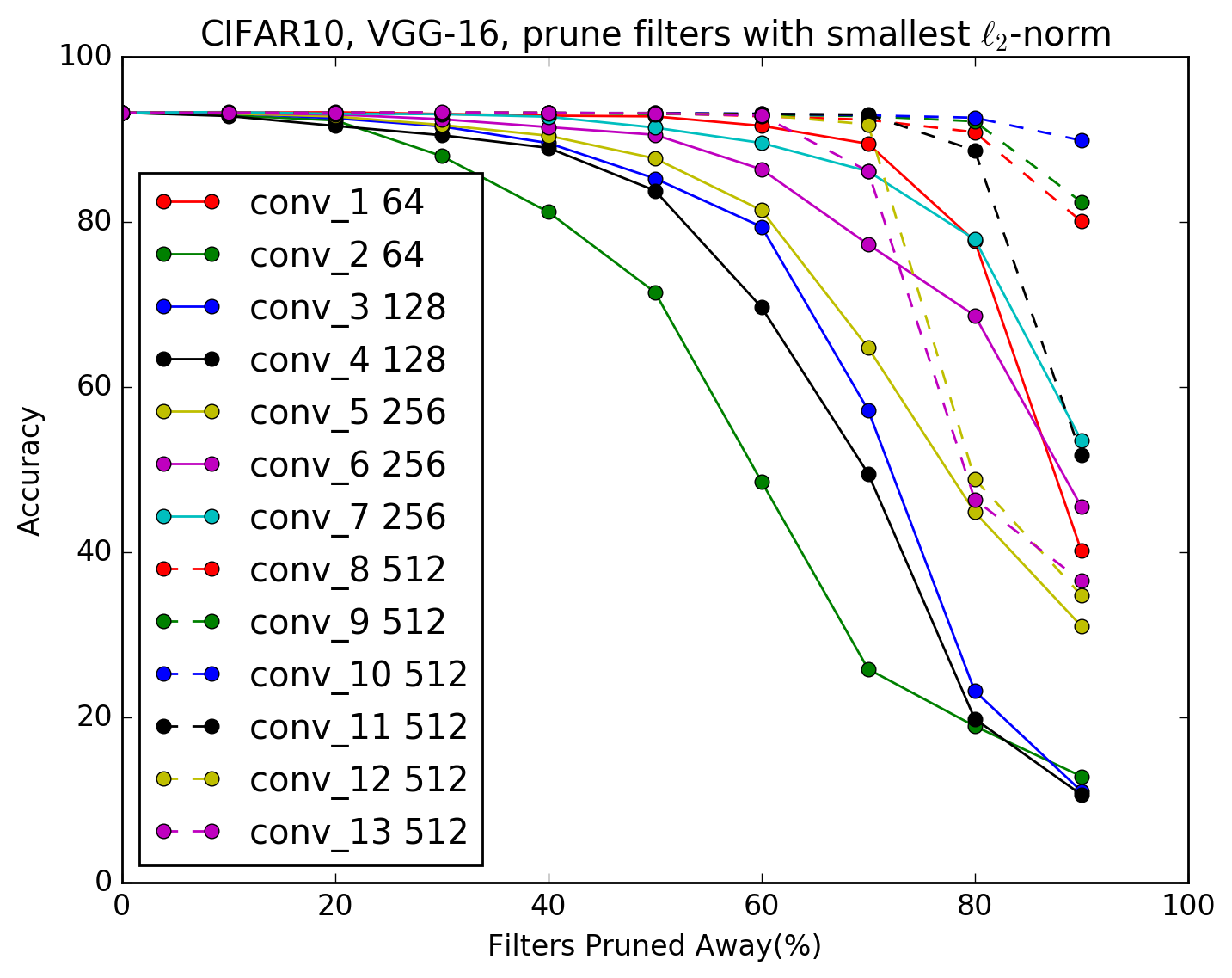}
   }
\end{tabular}
\caption{Comparison of $\ell_1$-norm and $\ell_2$-norm based filter pruning for VGG-16 on CIFAR-10.}
\label{fig:l2}
\end{figure*}

We compare $\ell_1$-norm with $\ell_2$-norm for filter pruning.
As shown in Figure \ref{fig:l2}, $\ell_1$-norm works slightly better than $\ell_2$-norm for layer $\texttt{conv\_2}$.
There is no significant difference between the two norms for other layers.

\subsection{FLOP and Wall-Clock Time}
FLOP is a commonly used measure to compare the computation complexities of CNNs.
It is easy to compute and can be done statically, which is independent of the underlying hardware and software implementations.
Since we physically prune the filters by creating a smaller model and then copy the weights, there are no masks or sparsity introduced to the original dense BLAS operations.
Therefore the FLOP and wall-clock time of the pruned model is the same as creating a model with smaller number of filters from scratch.

We report the inference time of the original model and the pruned model on the test set of CIFAR-10 and the validation set of ILSVRC 2012, which contains 10,000 $32\times32$ images and 50,000 $224\times224$ images respectively. 
The ILSVRC 2012 dataset is used only for ResNet-34.
The evaluation is conducted in Torch7 with Titan X (Pascal) GPU and cuDNN v5.1, using a mini-batch size 128.
As shown in Table~\ref{tab:wallclock_time}, the saved inference time is close to the FLOP reduction.
Note that the FLOP number only considers the operations in the Conv and FC layers, while some calculations such as Batch Normalization and other overheads are not accounted.

\begin{table}[htbp]
\centering
\small
\caption{The reduction of FLOP and wall-clock time for inference.}
\label{tab:wallclock_time}
\begin{tabular}{lllllrl}
\toprule
         & Model                          & FLOP                &  Pruned \%  & Time (s)  & Saved \% \\ \hline
         & VGG-16                         & $3.13 \times 10^8$  &             & 1.23      & \\
         & VGG-16-pruned-A                & $2.06 \times 10^8$  &  34.2\%     & 0.73      & 40.7\% \\ \hline
         & ResNet-56                      & $1.25 \times 10^8$  &             & 1.31      & \\         
         & ResNet-56-pruned-B             & $9.09 \times 10^7$  &  27.6\%     & 0.99      & 24.4\%\\ \hline
         & ResNet-110                     & $2.53 \times 10^8$  &             & 2.38      & \\         
         & ResNet-110-pruned-B            & $1.55 \times 10^8$  &  38.6\%     & 1.86      & 21.8\%\\ \hline      
         & ResNet-34                      & $3.64 \times 10^9$  &             & 36.02     & \\
         & ResNet-34-pruned-B             & $2.76 \times 10^9$  &  24.2\%     & 22.93     & 28.0\%\\ \bottomrule
\end{tabular}
\end{table}

\end{document}